\documentclass{article} 
\usepackage[final]{colm2026_conference}

\usepackage{microtype}
\usepackage{hyperref}
\usepackage{url}
\usepackage{booktabs}

\usepackage{lineno}

\definecolor{darkblue}{rgb}{0, 0, 0.5}
\hypersetup{colorlinks=true, citecolor=darkblue, linkcolor=darkblue, urlcolor=darkblue}

\usepackage{graphicx}
\usepackage{amsmath}
\usepackage{subcaption}
\usepackage{wrapfig}
\usepackage{booktabs}
\usepackage[dvipsnames]{xcolor}
\usepackage{soul}
\usepackage{pifont}

\newcommand{\cmark}{\ding{51}} 
\newcommand{\xmark}{\ding{55}} 

\newcommand{\highlight}[2]{\sethlcolor{#1}\hl{#2}}
\definecolor{cyan10}{HTML}{E5F6FF}
\definecolor{cyan20}{HTML}{BAE6FF}
\definecolor{cyan60}{HTML}{0072c3}
\definecolor{cyan70}{HTML}{00539a}
\definecolor{cyan80}{HTML}{003a6d}
\definecolor{teal10}{HTML}{D9FBFB}
\definecolor{teal20}{HTML}{9EF0F0}
\definecolor{teal60}{HTML}{007d79}
\definecolor{orange10}{HTML}{FFF2E8}
\definecolor{orange20}{HTML}{FFD9BE}
\definecolor{orange60}{HTML}{ba4e00}
\definecolor{blue10}{HTML}{EDF5FF}
\definecolor{blue20}{HTML}{D0E2FF}
\definecolor{blue70}{HTML}{0043ce}
\definecolor{blue80}{HTML}{002d9c}
\definecolor{magenta10}{HTML}{FFF0F7}
\definecolor{magenta20}{HTML}{FFD6E8}
\definecolor{magenta30}{HTML}{ffafd2}
\definecolor{magenta50}{HTML}{ee5396}
\definecolor{magenta60}{HTML}{d02670}
\definecolor{magenta70}{HTML}{9f1853}
\definecolor{purple10}{HTML}{F6F2FF}
\definecolor{purple20}{HTML}{E8DAFF}
\definecolor{purple30}{HTML}{d4bbff}
\definecolor{purple70}{HTML}{8a3ffc}
\definecolor{rose10}{HTML}{FCF2ED}
\definecolor{rose20}{HTML}{F9D9D1}
\definecolor{rose60}{HTML}{ab5638}
\definecolor{rose70}{HTML}{853c27}
\definecolor{red10}{HTML}{FFF1F1}
\definecolor{red20}{HTML}{FFD7D9}
\definecolor{green10}{HTML}{DEFBE6}
\definecolor{green20}{HTML}{A7F0BA}
\definecolor{green70}{HTML}{0e6027}
\definecolor{green80}{HTML}{044317}
\definecolor{yellow10}{HTML}{fcf4d6}
\definecolor{yellow20}{HTML}{fddc69}
\definecolor{gray20}{HTML}{e0e0e0}
\definecolor{gray30}{HTML}{c6c6c6}
\definecolor{gray40}{HTML}{a8a8a8}
\definecolor{gray80}{HTML}{393939}

\newenvironment{itemize*}%
 {\leftmargini=10pt\begin{itemize}%
  \setlength{\itemsep}{0pt}%
  \setlength{\parskip}{0pt}%
  }%
 {\end{itemize}}
\newenvironment{enumerate*}%
 {\begin{enumerate}%
  \setlength{\itemsep}{0pt}%
  \setlength{\parskip}{0pt}}%
 {\end{enumerate}}

\usepackage[most,skins,theorems]{tcolorbox}
\tcbset{
  aibox/.style={
    width=\textwidth,
    top=7pt,
    bottom=2pt,
    colback=blue!6!white,
    colframe=black,
    colbacktitle=black,
    enhanced,
    center,
    attach boxed title to top left={yshift=-0.1in,xshift=0.15in},
    boxed title style={boxrule=0pt,colframe=white,},
  }
}
\newtcolorbox{AIbox}[2][]{aibox,title=#2,#1}

\title{When Do LLMs Admit Their Mistakes? \\ Understanding The Role Of Model Belief In Retraction}

\author{%
Yuqing Yang\\
University of Southern California \\
\texttt{yyang063@usc.edu}
\And Robin Jia\\
University of Southern California \\
\texttt{robinjia@usc.edu}
}

\begin{document}

\ifcolmsubmission
\linenumbers
\fi

\maketitle

\begin{abstract}
We study the internal mechanisms that govern when LLMs choose to retract wrong answers, i.e., spontaneously and immediately acknowledge errors in their previously generated false assertions. Using model-specific testbeds, we find that while LLMs are capable of retraction, they do so only rarely, even when they can recognize their mistakes when asked in a separate interaction. We identify a reliable predictor of retraction: the model's \emph{momentary belief}, as measured by a linear probe on its internal representation. The probe is trained to predict the correctness of answers on external datasets unrelated to retraction, then applied to settings where models should retract. A model retracts only when it ``believes'' its answers to be incorrect \emph{during generation}; these beliefs frequently diverge from models' parametric knowledge as measured by factoid questions. Steering experiments further demonstrate that model belief causally drives retraction. In particular, when the model believes its answer to be incorrect, this not only encourages the model to attempt further verification, but also alters attention dynamics to promote retraction. Finally, we show that supervised fine-tuning re-uses this existing mechanism linking belief with retraction, and primarily improves retraction performance by helping the model learn more accurate internal beliefs.
\end{abstract}

\section{Introduction}

\begin{wrapfigure}{r}{0.49\textwidth}
    \vspace{-\intextsep}
    \centering
    \begin{minipage}{\linewidth}
    \centering
    \includegraphics[width=0.85\linewidth]{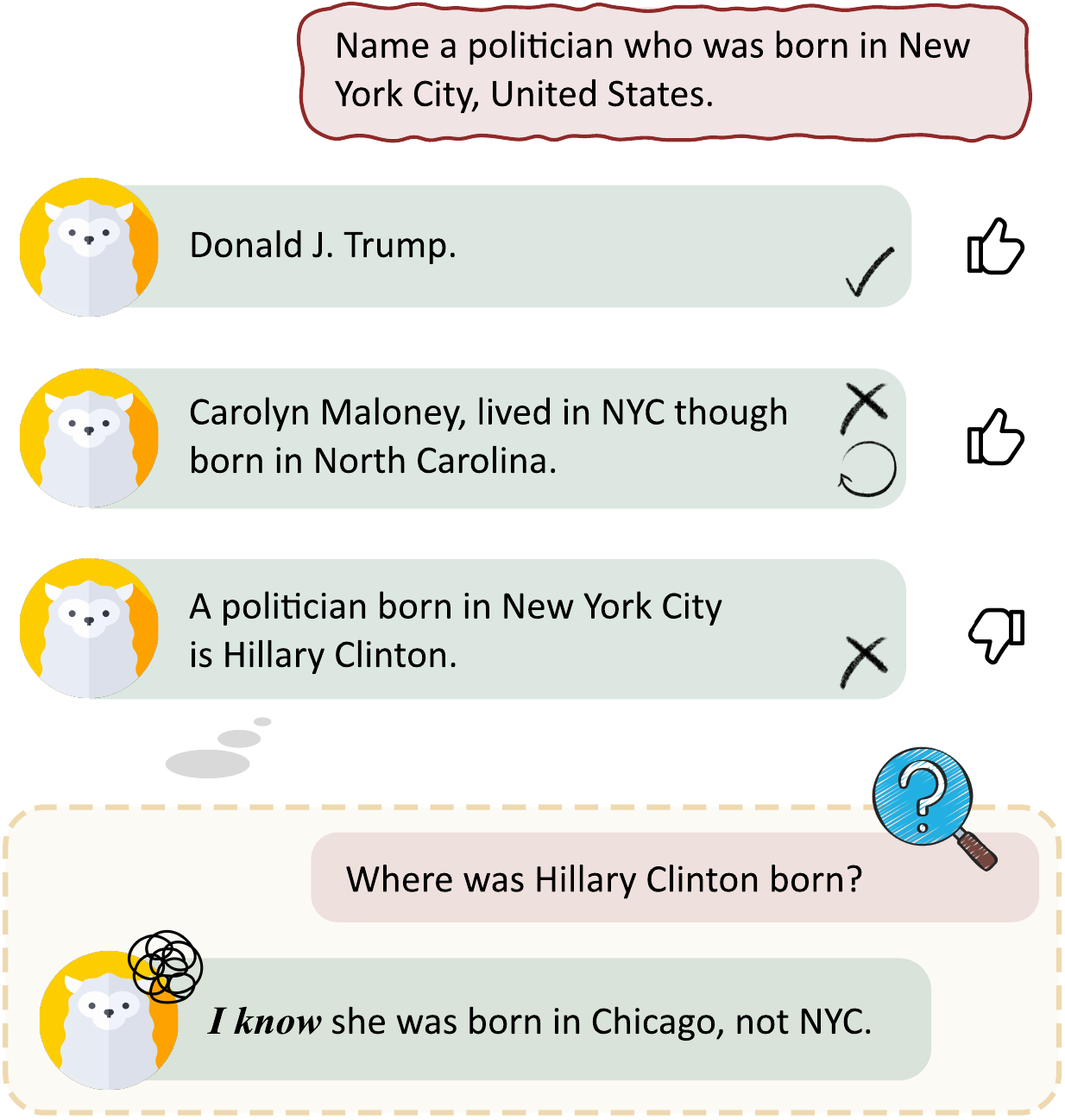}
    \vspace{-1mm}
    \caption{\small
\raisebox{-.2ex}{\includegraphics[height=2ex]{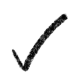}} correct,
\raisebox{-.2ex}{\includegraphics[height=2ex]{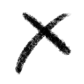}} wrong,
\raisebox{-.2ex}{\includegraphics[height=2ex]{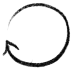}} retraction.
We study when LLMs fail to retract despite knowing the answer is wrong in verification questions.}
    \label{fig:intro}
    \end{minipage}
    \vspace{-5mm}
\end{wrapfigure}

Despite rapid progress, hallucinations \citep{DBLP:journals/corr/abs-2309-01219,kalai2025languagemodelshallucinate} remain a fundamental challenge for current large language models (LLMs), even when they appear to have relevant parametric knowledge \citep{DBLP:conf/icml/ZhangPMLS24,DBLP:conf/naacl/JiangQHFCMYZZ24,DBLP:journals/corr/abs-2410-22071}. Beyond preventing such potentially correctable errors outright \citep{DBLP:conf/nips/0002PVPW23,DBLP:journals/corr/abs-2310-01405}, an alternative post-hoc remedy is when a model, after hallucinating, \textit{spontaneously} recognizes and acknowledges its mistake---an act we define as \textbf{\textit{retraction}}, as illustrated in Figure~\ref{fig:intro}. When done by an LLM without external prompting, retraction reduces the user's burden of interrogating the model, while mitigating misinformation risk and enhancing reliability. In this work, we focus on knowledge-related questions and investigate the internal mechanisms that govern when LLMs choose to retract their incorrect answers.\footnote{We do not study large reasoning models (e.g., \citealp{DBLP:journals/corr/abs-2501-12948}), which frequently retract (e.g., ``Wait, no, that's not right...'') but also redundantly explore alternative answers \citep{DBLP:journals/corr/abs-2412-21187}. Instead, we focus on standard LLMs, aiming to inspire research on mitigating hallucinations through retraction when fast answers are desired.}

To study retraction, we first build model-specific testbeds called continuation datasets. In these datasets, we first prompt a model to produce an answer, then have it  \textit{continue} generating text to see whether it retracts its answer if it was incorrect. We leverage two datasets that frequently induce hallucinations: (1) \textsc{Wikidata}, which requires satisfying two conditions for correctness (e.g., \textit{``Name a \underline{politician} who was \underline{born in New York City}''};  \citealp{DBLP:conf/acl/DhuliawalaKXRLC24}), and (2) \textsc{Celebrity}, which asks for a celebrity given their lesser-known parents \citep{DBLP:conf/iclr/BerglundTKBSKE24}. For each question, we collect incorrect model answers and focus on the cases where the model's responses to verification questions (e.g., \textit{``Where was Hillary Clinton born?''}; \citealp{DBLP:conf/acl/DhuliawalaKXRLC24}) indicate that it knows the answer is incorrect; thus, in principle the model ``knows'' that it should retract. We find that models do sometimes retract their own incorrect answers, but they are generally reluctant to do so despite having the requisite knowledge.

This raises the question: \textit{Why do models fail to retract in these cases?} We consider two hypotheses: (1) Models may internally believe that their wrong answers are true, which causes them to not retract, or (2) Models may know that their answers are false, yet still choose not to verbalize this belief. By analyzing model internal states, we show that (1) is correct. We first leverage trained ``truthfulness'' probes from prior work that extract models' beliefs about whether a given statement is factually correct based on the current hidden state \citep{DBLP:conf/emnlp/AzariaM23,DBLP:conf/nips/0002PVPW23,DBLP:conf/emnlp/LiuCCH24}. On our datasets, we find that these probes cannot distinguish between correct and incorrect model-generated answers, implying that models ``believe'' their own incorrect answers during generation. Since our datasets only contain examples where models can recognize their answers as incorrect when asked separately, this result shows that their ``momentary'' beliefs during generation often contradict their own parametric knowledge. Meanwhile, these same probes are much better indicators of whether the model will retract: models tend to retract answers they internally believe are wrong and commit to those they believe are correct. Taken together, these findings establish that ``truthfulness'' indicators from prior work correlate with model retraction behavior, and incorrect beliefs about truthfulness can explain models' failures to retract wrong answers.

We further show that this link is causal: We can directly alter the model's retraction behavior by intervening on this belief direction.
Steering the model to believe an answer is factually correct (positive belief steering) suppresses retraction, while steering it to believe an answer is incorrect (negative belief steering) strongly promotes retraction. 
By analyzing the steered models, we identify two separate pathways through which internal beliefs control retraction. Negative belief steering first encourages the model to generate additional information (e.g., the entity's birthplace) for verification rather than stopping immediately after the answer. It then increases the model's attention to answer tokens and refines their attention value vectors, which further promotes retraction.

Finally, we show that supervised fine-tuning (SFT) leverages this same connection between the model's belief and retraction to improve model behavior.
We fine-tune the model to retract wrong answers and commit to correct ones; the model learns to adopt this desired behavior on in-distribution test data, consistent with prior work \citep{DBLP:conf/iclr/PrakashSHBB24,DBLP:journals/corr/abs-2408-16293,DBLP:journals/corr/abs-2501-19393}. 
We additionally find that the original belief direction continues to regulate retraction behavior. By probing the fine-tuned models, we show that SFT works by aligning the model's internal belief more closely with factual correctness, leading to more accurate retraction decisions. This bridges mechanistic interpretability with training-based approaches, strengthening the robustness and generality of our findings.

To summarize, our contributions are as follows: (1) We construct model-specific testbeds to evaluate an LLM's retraction performance, and show that current LLMs can retract but do so only rarely. (2) We uncover a connection between a model's internal belief and its external retraction behavior, and identify the underlying mechanism that governs this behavior. (3) We demonstrate that the causal influence of internal beliefs on retraction generalizes to supervised fine-tuned models, where more accurate beliefs lead to improved retraction performance. Code and data are available at \url{https://github.com/ayyyq/llm-retraction}.

\section{Related Work}

\subsection{Self-Correction in LLMs}

A closely related concept to retraction is self-correction. Retraction can be viewed as an important step within self-correction but does not require producing a correct final answer as the goal. Previous work on self-correction primarily relies on multi-turn procedures, such as asking the model verification questions \citep{DBLP:conf/acl/DhuliawalaKXRLC24,DBLP:journals/corr/abs-2405-14092}, prompting it to give feedback \citep{DBLP:conf/nips/MadaanTGHGW0DPY23,DBLP:journals/corr/abs-2412-14959,DBLP:conf/emnlp/LiuG0HZQZ23}, or directly instructing it to verify its initial responses \citep{DBLP:journals/corr/abs-2207-05221,DBLP:journals/corr/abs-2412-19513}. By contrast, we study retraction as a \textit{spontaneous} and \textit{immediate} behavior, happening without explicit prompts to identify errors. This distinction is practically important, as users may not ask an LLM to re-check its answers. Relatively fewer studies have examined spontaneous self-correction \citep{DBLP:journals/corr/abs-2408-16293,DBLP:journals/corr/abs-2506-06923}, showing that it can be acquired through elaborate training. Our work differs in that we explain how retraction emerges from the model's internal representations, complementing training-based approaches.

\subsection{Probing LLMs' Beliefs}

A series of studies leverages LLM's internal representations to probe for truthfulness \citep{DBLP:conf/emnlp/AzariaM23,DBLP:journals/corr/abs-2310-06824,DBLP:conf/nips/0002PVPW23,DBLP:conf/emnlp/LiuCCH24}. For example, \citet{DBLP:conf/emnlp/LiuCCH24} propose the existence of a universal ``truthfulness hyperplane'' that separates true and false statements by training on a diverse collection of true-false datasets. However, many works \citep{DBLP:conf/nips/0002PVPW23,DBLP:conf/emnlp/LiuCCH24} evaluate probes only on synthetically constructed true-false claims, where they achieve high performance, but such settings may not reflect the distribution of hallucinations in real LLM outputs. Indeed, while some research demonstrates strong performance in detecting hallucinations on in-distribution, model-generated data \citep{DBLP:conf/emnlp/AzariaM23,DBLP:conf/acl/CH-WangDEK24,DBLP:journals/corr/abs-2410-02707}, these approaches often fail to generalize to out-of-distribution examples \citep{DBLP:journals/corr/abs-2307-00175,DBLP:conf/acl/ServedioBPAN25}.

In line with prior work, we train probes on external true-false datasets but evaluate them on the model's own generated answers. Our findings provide a possible explanation for their limited generalization: such probes may not directly capture truth per se, but rather a model's \textit{internal belief}---its own judgments about the truth of the world \citep{DBLP:journals/corr/abs-2307-00175,DBLP:journals/corr/abs-2404-18865} during generation, which can diverge from ground-truth correctness. Crucially, we show that these belief signals are predictive of retraction behavior, suggesting that these probes may be tapping into dimensions of error awareness rather than factual truth itself.

\section{Task Definition and Preliminary Results}

\subsection{Task Definition}
Retraction denotes a model's immediate acknowledgment that its generated answer is incorrect or does not fully satisfy the user's requirements, regardless of whether it later produces a correct answer. To evaluate the retraction performance of current LLMs, we construct model-specific testbeds. We first collect questions from two knowledge-related datasets, \textsc{Wikidata} (e.g., ``Name a writer who was born in Oran, Algeria'') and \textsc{Celebrity} (e.g., ``Name a child of Joe Jackson''), which tend to elicit wrong answers, thereby creating a great opportunity to study retraction. Details of these two original datasets are provided in Appendix~\ref{sec:app-raw-dataset}.

\paragraph{Continuation Dataset.} 
Based on the collected questions, we construct model-specific continuation datasets. Each example pairs a question with a model-generated answer, after which the model is prompted to \textit{continue} generating to test whether it will retract, as illustrated below:

\begin{quote}
\textsc{USER}: Name a politician who was born in New York City.

\textsc{ASSISTANT}: \textcolor{orange}{Hillary Clinton}\textit{[Model generation continues from here...]}
\end{quote}

To ensure that each incorrect answer is, in principle, correctable by the tested LLM, we first sample answers from the model via temperature decoding. For each answer, we create verification questions (e.g., ``Where was \textcolor{orange}{\{model's answer\}} born?''; ``What is \textcolor{orange}{\{model's answer\}}'s profession?'') and check whether the model's responses to these questions conflict with the requirements of the original question, inspired by \citet{DBLP:conf/acl/DhuliawalaKXRLC24}. We retain two types of examples:
\begin{itemize*}
    \item \textbf{Correct Examples}: The answer is factually correct, and the model can correctly answer all verification questions.
    \item \textbf{Wrong Examples}: The answer is factually incorrect, and the model's responses to the verification questions contradict the original question, indicating that it should know the answer is incorrect.
\end{itemize*}

\begin{wraptable}{r}{0.5\textwidth}
    \centering
    \vspace{-\intextsep}
    \resizebox{0.5\textwidth}{!}{
    \begin{tabular}{lcccc}
        \toprule
         & \multicolumn{2}{c}{\textbf{\textsc{Wikidata}}} & \multicolumn{2}{c}{\textbf{\textsc{Celebrity}}} \\
         \cmidrule(lr){2-3}\cmidrule(lr){4-5}
         & \# Train & \# Test & \# Train & \# Test \\
        \midrule
        Llama3.1-8B & 1934 & 1202 & 1550 & 826 \\
        Qwen2.5-7B & 1496 & 1072 & -- & 1142 \\
        Olmo2-7B & 1796 & 1260 & -- & 1209 \\
        \bottomrule
    \end{tabular}}
    \caption{\small Continuation dataset statistics. Note that Qwen2.5-7B and Olmo2-7B have no \textsc{Celebrity} training sets due to too few correct examples, which are used only for SFT in Section~\ref{sec:sft}.}
    \label{tab:data_stat}
\end{wraptable}
 
We experiment with three popular LLMs from different model families, Llama3.1-8B-Instruct (\citealp{DBLP:journals/corr/abs-2407-21783}, abbr. Llama3.1-8B), Qwen2.5-7B-Instruct (\citealp{DBLP:journals/corr/abs-2412-15115}, abbr. Qwen2.5-7B), and Olmo2-1124-7B-Instruct (\citealp{DBLP:journals/corr/abs-2501-00656}, abbr. Olmo2-7B). The data statistics are listed in Table~\ref{tab:data_stat}. Details of the continuation datasets are provided in Appendix~\ref{sec:app-dataset}. In the following sections, we use \textsc{Wikidata} and \textsc{Celebrity} to denote the model-specific continuation datasets instead of the original datasets.

\paragraph{Evaluation Metrics.} We use Llama3.3-70B-Instruct as a judge \citep{DBLP:conf/nips/ZhengC00WZL0LXZ23} to automatically assess whether the tested model retracts the given answer in its response. The judge agrees with human annotations on 39/40 \textsc{Wikidata} and 38/40 \textsc{Celebrity} examples, and with an independent GPT-4.1-mini judge at $\kappa > 0.93$ on both datasets; see Appendix~\ref{sec:llm-as-a-judge} for details. We then calculate the following two metrics to evaluate the model's retraction performance:
\[\text{Retraction Recall} = \frac{|\text{Wrong \& Retraction}|}{|\text{Wrong}|},\,
\text{Retraction Precision} = \frac{|\text{Wrong \& Retraction}|}{|\text{Retraction}|}.\]
$|\text{Wrong}|$ denotes the number of wrong examples, and $|\text{Retraction}|$ indicates the number of examples that the tested model retracts according to the judgment of Llama3.3-70B-Instruct. Higher retraction recall and precision together represent better retraction performance.

\subsection{Models Can Retract, but Do So Infrequently}

\begin{wraptable}{r}{0.5\textwidth}
    \centering
    \vspace{-\intextsep}
    \resizebox{0.5\textwidth}{!}{
    \begin{tabular}{lcccc}
        \toprule
         & \multicolumn{2}{c}{\textbf{\textsc{Wikidata}}} & \multicolumn{2}{c}{\textbf{\textsc{Celebrity}}} \\
         \cmidrule(lr){2-3}\cmidrule(lr){4-5}
         & Prec. & Rec. & Prec. & Rec. \\
        \midrule
        Llama3.1-8B & 0.9012 & 0.2579 & 0.7722 & 0.1477 \\
        Qwen2.5-7B & 0.8824 & 0.1119 & 0.9667 & 0.0290 \\
        Olmo2-7B & 0.9881 & 0.1317 & 0.8824 & 0.0150 \\
        \bottomrule
    \end{tabular}}
    \caption{\small Retraction performance on the \textsc{Wikidata} and \textsc{Celebrity} test sets across different LLMs.}
    \label{tab:preliminary_results}
\end{wraptable}

As shown in Table~\ref{tab:preliminary_results}, models consistently exhibit low but non-zero retraction recall on our datasets. We infer that LLMs have the capability to retract incorrect answers, but the consistently low recall (at most $25\%$) highlights that such retractions are rare.
Recall that our verification questions provide clear evidence that the model knows that the incorrect answers in our datasets are indeed incorrect. 
Thus, the model appears to have both the knowledge and the ability to retract. Then, why do LLMs fail to retract more incorrect answers? What factors govern their retraction behavior?

\section{Model Belief Guides Retraction}

\subsection{Probing for  Belief}
\label{sec:probe}
To investigate the gap between LLMs' parametric knowledge measured by factoid questions and their failure to retract incorrect answers, we build on prior work that probes internal representations of truthfulness \citep{DBLP:conf/emnlp/AzariaM23,DBLP:journals/corr/abs-2310-06824,DBLP:conf/nips/0002PVPW23}. Here, we use the term \textit{internal belief} rather than truthfulness to emphasize the distinction between a model's internal assessment of correctness and ground-truth correctness. Our key question is: when the model's parametric knowledge implies that its answer is wrong, 
does its internal representation reflect this during answer generation?

\paragraph{Universal Truthfulness Dataset.} To prevent overfitting to a single dataset, we follow \citet{DBLP:conf/emnlp/LiuCCH24} and train our probes on a diverse set of external true-false datasets, including 800 examples each from Natural Questions \citep{DBLP:journals/tacl/KwiatkowskiPRCP19}, Trivia QA \citep{DBLP:conf/acl/JoshiCWZ17}, and SciQ \citep{DBLP:conf/aclnut/WelblLG17}. All three are short-answer, closed-book QA tasks with a format similar to \textsc{Wikidata} and \textsc{Celebrity}. Each dataset is balanced with a 50/50 split of correct and incorrect answers, where the incorrect answers are generated by GPT-4-turbo. We denote this collection as Universal Truthfulness QA (UTQA) dataset.

\begin{figure*}[t]
  \centering
  \begin{subfigure}[t]{0.49\textwidth}
    \centering
    \includegraphics[width=\textwidth]{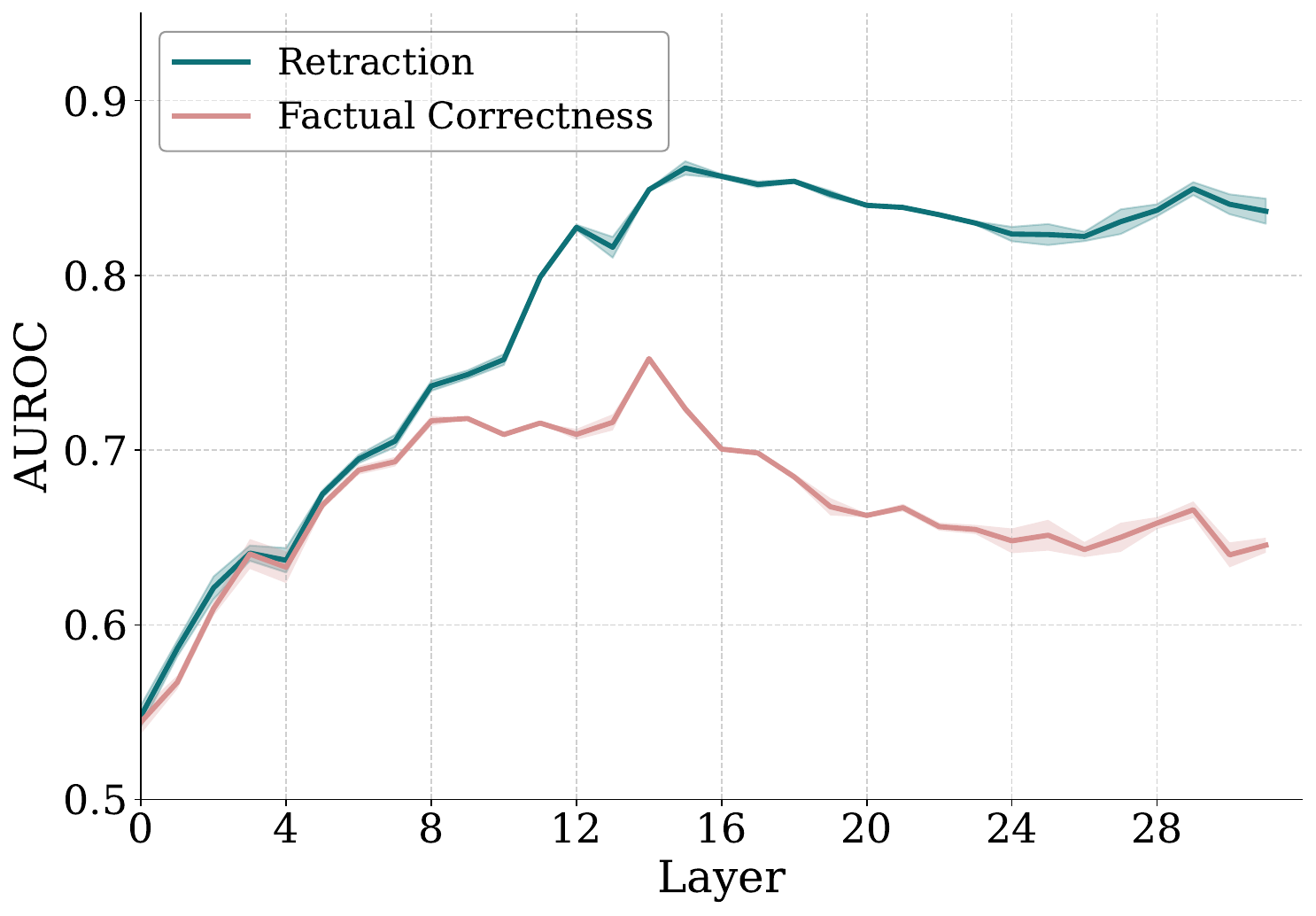}
    \caption{\small Llama3.1-8B on \textsc{Wikidata}.}
  \end{subfigure}
  \hfill
  \begin{subfigure}[t]{0.49\textwidth}
    \centering
    \includegraphics[width=\textwidth]{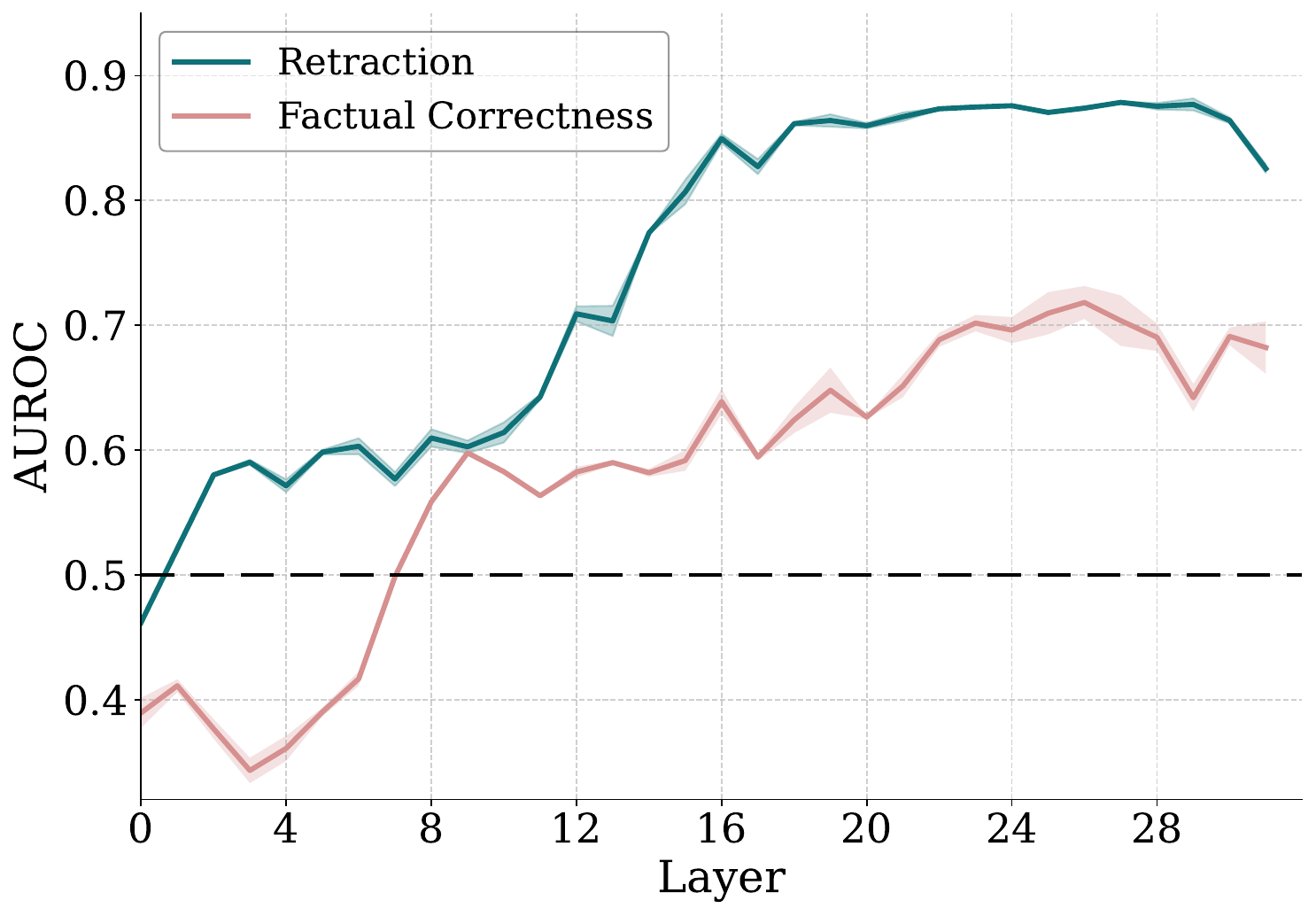}
    \caption{\small Llama3.1-8B on \textsc{Celebrity}.}
  \end{subfigure}
  \caption{\small Layer-wise AUROC of belief scores for predicting factual correctness and retraction in Llama3.1-8B. An AUROC of 0.5 corresponds to random guessing. Results are averaged over three runs with different random seeds, and error bars denote standard deviation.}
  \label{fig:probe_llama}
\end{figure*}

\paragraph{Probe Setup.} For each LLM layer, we train a linear probe on the UTQA dataset using the hidden states after the given answer. These probes learn to distinguish correct and incorrect answers on UTQA and thus serve as proxies for the model's internal belief. We then apply the probes to \textsc{Wikidata} and \textsc{Celebrity} examples to investigate how the model's internal belief relates to both factual correctness and retraction behavior. To quantify the relationship, we report AUROC (Area Under the Receiver Operating Characteristic Curve), treating belief scores as the decision score and either binary factual correctness or retraction labels as the ground truth. Because internal belief and retraction are hypothesized to be negatively correlated, we use $1 - \text{belief score}$ when predicting retraction. A higher AUROC indicates that belief scores are reliable predictors of the target label, while an AUROC of 0.5 implies no discriminative power.

\paragraph{Results.} From Figure~\ref{fig:probe_llama}, we can see that belief probes are less predictive of factual correctness but much more reliable for predicting retraction.

(1) Since factual correctness on our test sets reflects the model's parametric knowledge as assessed by verification questions, the suboptimal AUROC of belief scores indicates a misalignment between the model's momentary internal belief and its stored knowledge. This further verifies LLMs' limitations in knowledge manipulation \citep{DBLP:conf/iclr/Allen-ZhuL25a,DBLP:conf/iclr/BerglundTKBSKE24}, from the perspective of internal representations.

(2) More importantly, we find that \textbf{the model's internal belief, although obtained without retraction-related data, is a better indicator of whether the model retracts its own generated answers} than of factual correctness, with predictive power emerging in the middle layers. In particular, low belief scores correspond to retraction.\footnote{Internal belief is conceptually distinct from uncertainty. Appendix~\ref{sec:uncertainty} shows that uncertainty is a much weaker predictor of retraction.} This suggests that the probed direction captures the model's internal judgment at that moment instead of objective truth, and is manifested in its subsequent behavior (i.e., retracting or committing).

These results generalize across model families, including Qwen2.5-7B and Olmo2-7B, and extend to larger models such as Llama3.1-70B-Instruct, as shown in Appendix~\ref{sec:app-probe-plots}.

\subsection{Steering Internal Belief Affects Retraction}
\label{sec:steering}

Our probing results establish a correlation between the model's internal belief and its retraction behavior. To demonstrate that internal belief causally influences retraction behavior, we steer the model's hidden states towards positive belief (i.e., believe an answer is correct) and negative belief (i.e., believe an answer is incorrect) directions.

\paragraph{Activation Steering.} We still use the UTQA dataset to find steering directions. For each layer $l \in |L|$, we calculate the mean hidden states $h^+_l$ for correct answers at the last token of the answer, and $h^-_l$ for incorrect answers. We then compute the \textit{difference-in-means} vector $v_l = h^+_l - h^-_l$ \citep{difference-in-means,DBLP:conf/nips/ArditiOSPPGN24,DBLP:conf/nips/0002PVPW23}, which represents a linear belief direction. We add or subtract this difference-in-means vector to the activations of a new answer, thereby shifting the model's perception of the correctness of the answer: $h_l' \leftarrow h_l \pm \alpha v_l$, where $\alpha$ controls the strength of steering. Note that we steer only at the last token of the answer; we do not add the steering vector at any following generation steps in order to minimize disruption to the model's natural generation. Similar to prior work \citep{DBLP:journals/corr/abs-2308-10248,DBLP:conf/iclr/LeePRMDND25,DBLP:conf/nips/0002PVPW23}, we manually search for the steering hyperparameters to ensure that the steering is effective and minimally invasive, as detailed in Appendix~\ref{sec:hyperparams}.

\begin{figure*}[t]
  \centering
  \includegraphics[width=\textwidth]{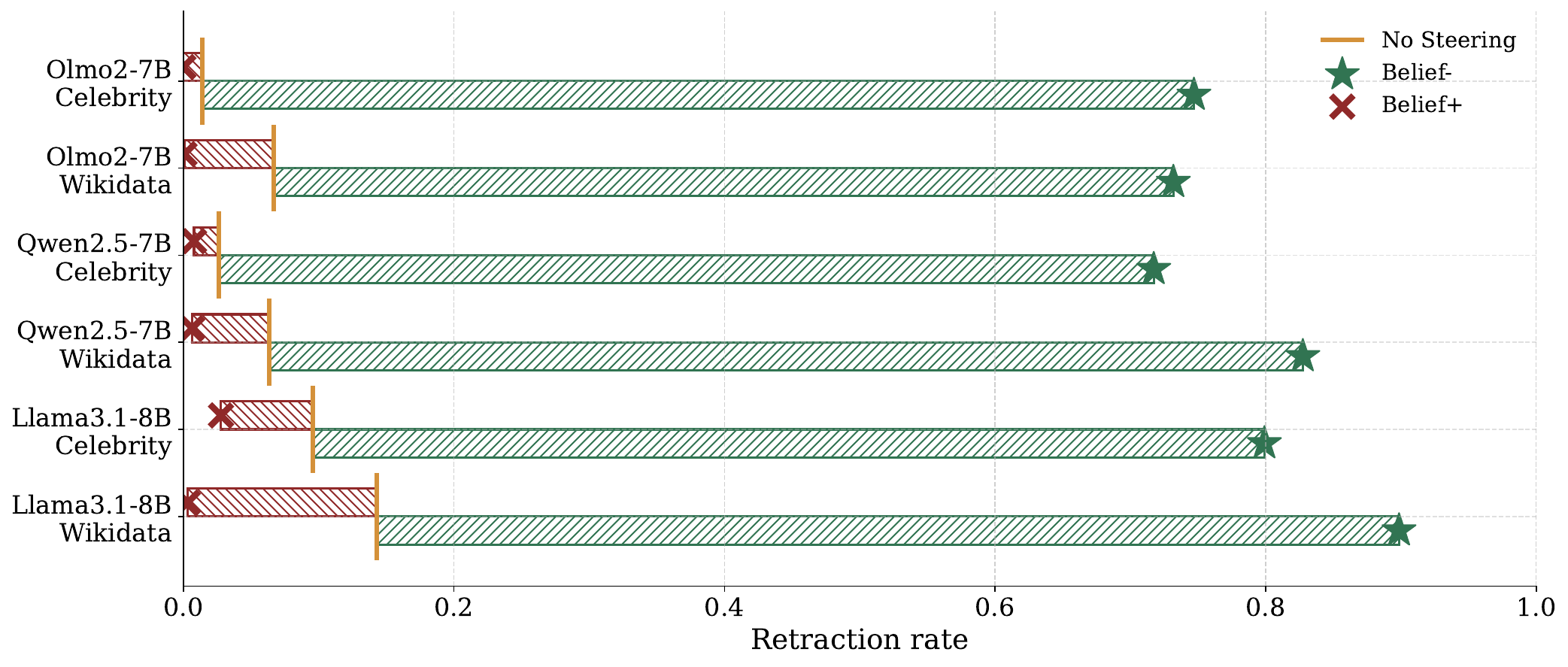}
  \caption{\small Retraction rate under belief steering. ``Belief-'' denotes negative belief steering while ``Belief+'' denotes positive belief steering.}
  \label{fig:interv_results}
\end{figure*}

\vspace{-2mm}
\paragraph{Results.} We present retraction rate (i.e., the proportion of retracted examples) in Figure~\ref{fig:interv_results} for clarity and provide detailed retraction recall and precision in Appendix Table~\ref{tab:llama_interv_f1}, ~\ref{tab:qwen_interv_f1}, and~\ref{tab:olmo_interv_f1}. From Figure~\ref{fig:interv_results}, we can see that across all three models and two datasets, belief steering effectively controls retraction behavior in both directions. Specifically, strengthening the model's belief in the negative direction causes it to retract over 70\% of the time across the entire dataset. In contrast, when we strengthen the model's belief in the positive direction, retraction rate drops to nearly zero, indicating the model rarely retracts. This supports our hypothesis about the role of model belief in retraction: an LLM tends to take back an answer only when it internally believes it is incorrect; otherwise, it is likely to stand by its initial answer.

We note that the belief steering direction is derived from out-of-distribution data, yet remains effective at modulating retraction behavior across different datasets and under variations in prompting and decoding hyperparameters (Appendix~\ref{sec:robust}). This contrasts with common steering approaches, which typically rely on in-distribution data (Appendix~\ref{sec:other_steer_direction}). More encouragingly, the causal link between belief and retraction extends beyond knowledge-based question answering. In GSM8k \citep{DBLP:journals/corr/abs-2110-14168}, a widely used math reasoning dataset, we identify cases where Llama3.1-8B makes an intermediate mistake. By applying negative belief steering at the token where the mistake happens, we induce the model to retract, leading it to ultimately get the correct answer in 20\% of these cases (Appendix~\ref{sec:math}).

Overall, we aim to understand when and why LLMs choose to retract. Taken together, both the probing and steering results support the conclusion that \textbf{the model's belief---defined independently of retraction and trained on separate data---causally affects retraction behavior and generalizes across different datasets.}

\section{Mechanistic Analysis}

Having established that retraction is guided by LLMs' internal beliefs, we now turn to a deeper investigation of how beliefs function. In this section, we explore the mechanisms through which beliefs shape model behavior, from surface-level token generation to deeper attention dynamics.

\subsection{Belief Influences the Decision to Stop Generating}
\label{sec:stop}

\begin{table}[ht]
    \centering
    \small
    \begin{tabular}{lcccccc}
        \toprule
         & \multicolumn{2}{c}{\textbf{Llama3.1-8B}} & \multicolumn{2}{c}{\textbf{Qwen2.5-7B}} & \multicolumn{2}{c}{\textbf{Olmo2-7B}} \\
         \cmidrule(lr){2-3}\cmidrule(lr){4-5}\cmidrule(lr){6-7}
         & \textsc{Wikidata} & \textsc{Celebrity} & \textsc{Wikidata} & \textsc{Celebrity} & \textsc{Wikidata} & \textsc{Celebrity} \\
        \midrule
        No Steering & 0.7413 & 0.6041 & 0.0028 & 0.0271 & 0.0563 & 0.1960 \\
        Belief- & 0.0017 & 0.0206 & 0.0271 & 0.0096 & 0.0000 & 0.0000 \\
        Belief+ & 0.9867 & 0.8765 & 0.4310 & 0.8126 & 0.9992 & 0.9992 \\
        \bottomrule
    \end{tabular}
    \caption{\small Stop rate: the fraction of examples where generation stops right after the given answer.}
    \label{tab:stop_rate}
\end{table}

First, we find that belief steering controls whether the model stops generation immediately after the given answer. If the model outputs a ``\texttt{.}'' or ``\texttt{EOS}'' token directly following the answer, we define this as a \textit{stop} and calculate the stop rate in Table~\ref{tab:stop_rate}.

\begin{wraptable}{r}{0.5\textwidth}
    \centering
    \vspace{-\intextsep}
    \resizebox{0.5\textwidth}{!}{
    \begin{tabular}{lcccc}
        \toprule
         & \multicolumn{2}{c}{\textbf{\textsc{Wikidata}}} & \multicolumn{2}{c}{\textbf{\textsc{Celebrity}}} \\
         \cmidrule(lr){2-3}\cmidrule(lr){4-5}
         & Prec. & Rec. & Prec. & Rec. \\
        \midrule
        No Steering & 0.9012 & 0.2579 & 0.7722 & 0.1477 \\
        \midrule
        \multicolumn{3}{l}{\textbf{\textit{Appending ``is''}}} \\
        No Steering & 0.8254 & 0.5740 & 0.8310 & 0.1429 \\
        Belief- & 0.5026 & 0.9717 & 0.4836 & 0.8232 \\
        Belief+ & 0.9847 & 0.3211 & 0.8108 & 0.0726 \\
        \bottomrule
    \end{tabular}}
    \caption{\small Retraction performance for Llama3.1-8B under the \textit{is}-appended setting.}
    \label{tab:llama_fixed_is}
\end{wraptable}

We observe that positive belief steering increases stop rate, suggesting that when the model believes the answer is true, it is more likely to terminate generation early, foregoing the opportunity to verify the answer. In contrast, negative belief steering reduces stop rate: the model tends to generate additional information like the entity's birthplace and profession, which encourages it to reflect on and potentially challenge its initial answer.

At the same time, belief steering does more than just change the immediate next token. To demonstrate this, we append ``is'' after the given answer to prevent early stopping, e.g., ``\textcolor{orange}{Hillary Clinton}\underline{ is}\textit{[Model generation continues from here...]}''. As shown in Table~\ref{tab:llama_fixed_is}, simply appending a continuation token can, in some cases, increase retraction recall for Llama3.1-8B, leading to improved retraction performance. Belief steering under this \textit{is}-appended setting still further increases retraction recall, indicating that its influence extends beyond influencing the immediate next token.

\subsection{Belief Influences Retraction Primarily via Attention Value Vectors}

Given that retraction behavior is closely tied to previous context and involves the model's attention mechanisms, we next examine how belief steering modifies attention outputs to influence retraction.

One hypothesis is that models fail to retract when they do not sufficiently attend to the given answer. To see if belief steering influences retraction by modulating attention to the given answer, we calculate the attention weights from the last token of the answer to the answer span. In Appendix, Table~\ref{tab:attn_weights} presents the average change in attention weights under different belief steering directions. Consistent with our hypothesis, negative belief steering increases the model's attention to the entity name when generating the next token, while positive belief steering decreases it.

\textbf{Attention values have stronger causal influence on retraction than attention weights.}

\begin{wraptable}{r}{0.5\textwidth}
    \centering
    \vspace{-\intextsep}
    \resizebox{0.5\textwidth}{!}{
    \begin{tabular}{lcccc}
        \toprule
         & \multicolumn{2}{c}{\textbf{\textsc{Wikidata}}} & \multicolumn{2}{c}{\textbf{\textsc{Celebrity}}} \\
         \cmidrule(lr){2-3}\cmidrule(lr){4-5}
         & Prec. & Rec. & Prec. & Rec. \\
        \midrule
        No Steer & 0.8254 & 0.5740 & 0.8310 & 0.1429 \\
        \midrule
        \multicolumn{5}{l}{\textbf{\textit{belief-}}} \\
        Patch W & 0.7694 & 0.5940 & 0.8228 & 0.1574 \\
        Patch V & 0.5069 & 0.9784 & 0.5055 & 0.5569 \\
        Full Steer & 0.5026 & 0.9717 & 0.4836 & 0.8232 \\
        \midrule
        \multicolumn{5}{l}{\textbf{\textit{belief+}}} \\
        Patch W & 0.8261 & 0.5691 & 0.8261 & 0.1380 \\
        Patch V & 0.9851 & 0.3311 & 0.7955 & 0.0847 \\
        Full Steer & 0.9847 & 0.3211 & 0.8108 & 0.0726 \\
        \bottomrule
    \end{tabular}}
    \caption{\small Patching results for Llama3.1-8B under the \textit{is}-appended setting.}
    \label{tab:attn_patch_llama_fixed_is}
\end{wraptable}

Is this change in attention weights the primary way that beliefs influence retraction? We conduct patching experiments \citep{DBLP:conf/nips/MengBAB22,DBLP:conf/emnlp/GevaBFG23} to answer this question. Instead of directly adding steering vectors to the hidden states of each layer, we selectively retain specific components, such as attention weights or attention value vectors, from the steered model, and patch them into an unsteered model. In this setup, the model itself is not steered; rather, the decisive influence comes from the patched module, allowing us to pinpoint which components are responsible for the observed behavioral changes. We experiment with patching attention weights from salient heads (i.e., heads whose attention to the answer changes significantly after steering), as well as attention value vectors at all layers, for the last token of the answer (Refer to Appendix~\ref{sec:app-patching} for implementation details). We present patching results for Llama3.1-8B under the \textit{is}-appended setting in Table~\ref{tab:attn_patch_llama_fixed_is}, to mitigate the effect of next-token prediction.

We can observe that patching attention weights (i.e., \textit{Patching W}) does have a measurable effect on retraction; however, this effect is relatively modest. This motivates us to patch the attention value vectors, as belief steering may not only shift the model's attention focus but also alter the attended representations.

\begin{figure*}[t]
  \centering
  \begin{subfigure}[t]{0.49\textwidth}
    \centering
    \includegraphics[width=\textwidth]{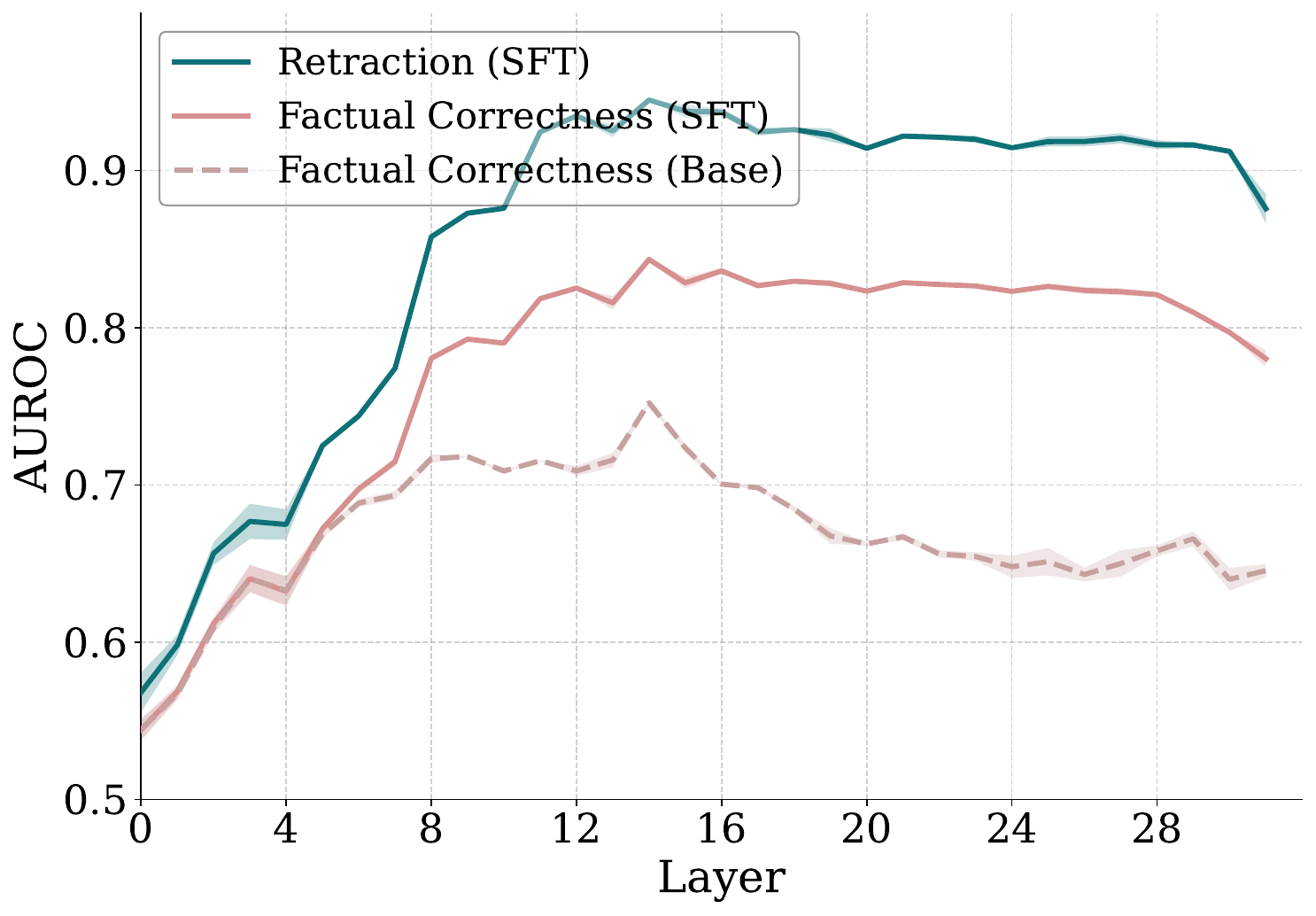}
    \caption{\small Llama3.1-8B on \textsc{Wikidata}.}
  \end{subfigure}
  \hfill
  \begin{subfigure}[t]{0.49\textwidth}
    \centering
    \includegraphics[width=\textwidth]{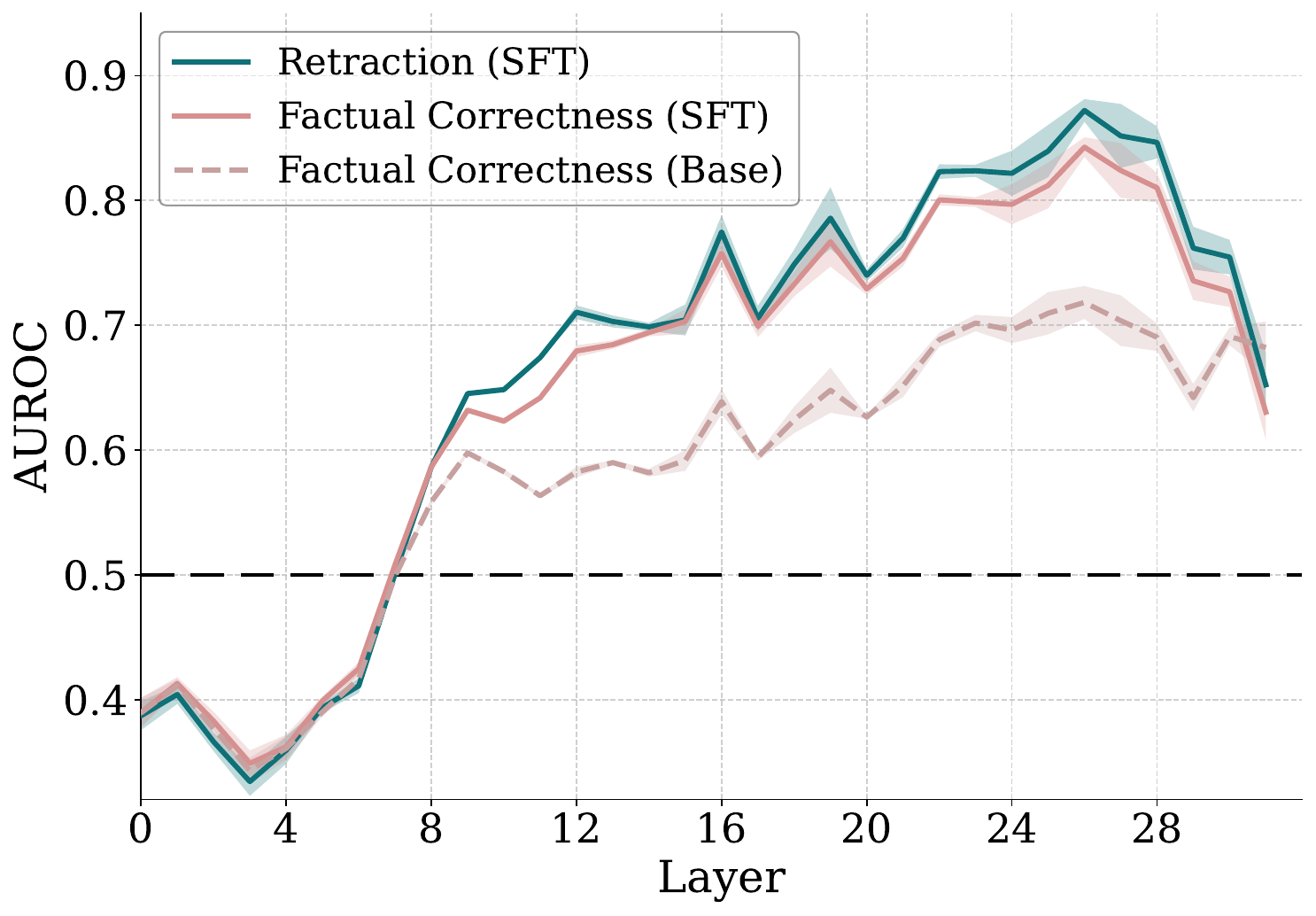}
    \caption{\small Llama3.1-8B on \textsc{Celebrity}.}
  \end{subfigure}
  \caption{\small Layer-wise AUROC of belief scores in Llama3.1-8B (Base) and its fine-tuned variant (SFT).}
  \label{fig:probe_sft_llama}
  \vspace{-4mm}
\end{figure*}

Compared to attention weights, patching attention value vectors (i.e., \textit{Patching V}) restores more the retraction behavior observed with full steering (i.e., \textit{Full Steer}) in both directions. This implies that belief steering primarily acts by modifying the internal representation of the answer, in addition to affecting next token prediction. This is further supported by experiments without appending \textit{is}, as well as by consistent results on Qwen2.5-7B and Olmo2-7B as shown in Appendix~\ref{sec:patch_attn_value}.

\section{Supervised Fine-Tuning Can Learn Better Internal Belief}
\label{sec:sft}
Given that SFT can enhance existing capabilities of LLMs \citep{DBLP:conf/iclr/PrakashSHBB24,DBLP:conf/nips/0004CQN024}, we first verify that straightforward SFT indeed improves in-distribution retraction performance. Specifically, we append ``is the correct answer.'' to correct examples and ``is not the correct answer.'' to wrong examples in the training data, thereby explicitly encouraging appropriate retraction behavior (training details can be found in Appendix~\ref{sec:app-sft-hp}). As shown in Table~\ref{tab:sft}, the fine-tuned model learns to distinguish factually correct from incorrect answers and to respond accordingly. Having established this, what remains underexplored is whether our findings on the role of model belief in retraction continue to hold after fine-tuning.

\begin{wraptable}{r}{0.5\textwidth}
    \centering
    \vspace{-\intextsep}
    \resizebox{0.5\textwidth}{!}{
    \begin{tabular}{lcccc}
        \toprule
         & \multicolumn{2}{c}{\textbf{\textsc{Wikidata}}} & \multicolumn{2}{c}{\textbf{\textsc{Celebrity}}} \\
         \cmidrule(lr){2-3}\cmidrule(lr){4-5}
         & Prec. & Rec. & Prec. & Rec. \\
        \midrule
        Baseline & 0.9012 & 0.2579 & 0.7722 & 0.1477 \\
        \midrule
        SFT & 0.7815 & 0.8453 & 0.8988 & 0.9031 \\
        Belief- for SFT & 0.5013 & 1.0000 & 0.5092 & 1.0000 \\
        Belief+ for SFT & 0.9144 & 0.2845 & 0.9407 & 0.5763 \\
        \bottomrule
    \end{tabular}}
    \caption{\small In-distribution supervised fine-tuning results and follow-up steering performance for LLaMA3.1-8B. Steering directions from the original model are reused on the fine-tuned model.}
    \label{tab:sft}
\end{wraptable}

We first apply the same belief steering vectors from the original model and the same hyperparameters\footnote{Note that these may not be the optimal hyperparameters. In fact, extending steering from layers 6-14 to 6-20 reduces retraction recall on Belief+ \textsc{Celebrity} set from 0.5763 to 0.2300.} to steer the fine-tuned model. As shown in Table~\ref{tab:sft}, the steering vectors can be generalized to the fine-tuned model and change its retraction behavior in both directions, without altering its response format learned during SFT, i.e., ``is (not) the correct answer''. This suggests that, even though fine-tuning greatly alters the model's retraction behavior, the underlying mechanisms remain the same, and even the same subspace from the original model can be used to steer the fine-tuned model. Similar results for Qwen2.5-7B and Olmo2-7B, presented in Appendix~\ref{sec:app-sft}, further confirm this observation.

We also probe the model's internal belief after SFT, as shown in Figure~\ref{fig:probe_sft_llama}. Since we reuse the probes from the original model without re-training, there might be some distribution shift. Nevertheless, these probes based on internal representations remain effective at predicting external retraction behavior. More interestingly, SFT aligns the model's internal belief more closely with factual correctness, as reflected by the higher AUROC for \textit{Factual Correctness (SFT)} compared to \textit{Factual Correctness (Base)}. This suggests that SFT enables LLMs to form more accurate internal beliefs.

\section{Discussion}
\label{sec:discussion}
Our results turn retraction from an opaque behavior into a signal that is both interpretable and controllable through the belief direction. Two implications follow.

\paragraph{Retraction can be controlled at inference time.} Steering along the belief direction offers a lightweight, inference-time alternative to SFT, where a separate probe judges whether an answer is factually correct and steering is applied only where it is needed. On Llama3.1-8B, this raises retraction recall from 0.26 to 0.74 at 0.81 precision, approaching SFT's F1 without fine-tuning the model (Appendix~\ref{sec:conditional-steering}).

\paragraph{Fine-tuning shapes internal belief, not surface format.} SFT improves retraction by aligning the model's internal belief with factual correctness, rather than teaching a surface format (Section~\ref{sec:sft}). This implies internal belief is malleable and shaped by the correctness signal in the training data, highlighting the importance of accurate correctness labels when fine-tuning.

\section{Conclusion}

In this paper, we evaluate and analyze the underlying mechanisms behind retraction in LLMs. Using our model-specific continuation datasets, we find that while current LLMs are capable of retracting their own incorrect answers, they do so infrequently. Through probing and steering experiments, we demonstrate that retraction is causally influenced by the model's internal belief: a model fails to retract an incorrect answer when it internally believes it is correct. We further show that beliefs guide retraction by affecting both the surface-level token predictions and deeper attention dynamics. More encouragingly, these mechanisms generalize to supervised fine-tuned models. We hope our work contributes to the development of more transparent and reliable LLMs.

\section*{Acknowledgments and Disclosure of Funding}

Sincere thanks to Ting-Yun Chang and everyone in Allegro Lab.
This work was supported in part by the National Science Foundation under Grant No. IIS-2403436. Any opinions, findings, and conclusions or recommendations expressed in this material are those of the author(s) and do not necessarily reflect the views of the National Science Foundation.

\section*{Disclosure of LLM Usage}

We use LLMs as a judge to evaluate retraction performance. We did not use LLMs to originate research ideas, write original content in the paper, or generate data or plots.




\bibliography{colm2026_conference}
\bibliographystyle{colm2026_conference}

\appendix
\newpage

\section{Limitations}

There are several limitations for future research. First, although different LLMs share the same overall retraction mechanism---being causally influenced by the model's internal belief---the specific layers where this influence is most pronounced vary across models. As shown in Appendix~\ref{sec:hyperparams}, steering at early to mid layers is effective for Llama3.1-8B and Qwen2.5-7B, whereas Olmo2-7B requires intervention at higher layers to elicit stronger retraction. These differences likely stem from variations in training recipes, including data and optimization strategies used.

Second, our analysis primarily focuses on short-answer, knowledge-based question answering tasks. One natural extension is to long-form generation, such as ``Name 15 politicians who were born in New York City''. As many self-correction studies target reasoning tasks, it would also be valuable to examine whether our findings generalize to that domain. While we demonstrate that belief steering generalizes to GSM8k in preliminary experiments (see Appendix~\ref{sec:math}), several challenges remain, including accurately identifying steering points and isolating the influence of earlier outputs on later ones.

\section{Experimental Details}
\subsection{Details of Original Datasets}
\label{sec:app-raw-dataset}

\begin{wraptable}{r}{0.5\textwidth}
    \centering
    \vspace{-\intextsep}
    \begin{tabular}{lcc}
        \toprule
         & \textbf{\# Train} & \textbf{\# Test} \\
        \midrule
        \textsc{Wikidata} & 2000 & 1160 \\
        \textsc{Celebrity} & 1584 & 800 \\
        \bottomrule
    \end{tabular}
    \caption{Number of questions.}
    \vspace{-2mm}
    \label{tab:dataset_stat}
\end{wraptable}

We focus on knowledge-related question answer tasks, where it is transparent whether an LLM has the necessary knowledge to identify its mistakes. To facilitate the study of retraction, we collect questions from \textsc{Wikidata} and \textsc{Celebrity}, which are easy to induce hallucinations. The number of questions in each split of the datasets is reported in Table~\ref{tab:dataset_stat}

\paragraph{\textsc{Wikidata}.} \textsc{Wikidata} was originally proposed by \citet{DBLP:conf/acl/DhuliawalaKXRLC24}, and is characterized by each question containing two constraints---profession and birthplace---both of which must be satisfied for the answer to be correct. This makes the task challenging for LLMs, resulting in relatively low accuracy \citep{DBLP:conf/iclr/YuksekgonulCJGN24}. However, the original dataset was not publicly released. To reconstruct it, we collect a set of popular professions and cities, and generate new questions by pairing them. We retain only those combinations for which a correct answer exists. For accuracy evaluation, we query the Wikidata API\footnote{https://query.Wikidata.org/}. An example question is:
\begin{quote}
\texttt{Name a writer who was born in Oran, Algeria.}
\end{quote}

\paragraph{\textsc{Celebrity}.} \textsc{Celebrity} was originally introduced by \citet{DBLP:conf/iclr/BerglundTKBSKE24}. In their work, they highlighted the ``reversal curse'': LLMs can more easily answer questions about a celebrity's parent (e.g., ``Who is Tom Cruise's mother?'') than the reverse (e.g., ``Who is Mary Lee Pfeiffer's son?'', where the correct answer is Tom Cruise). We focus on the reverse questions. However, in their evaluation, a model was prompted 10 times per question and considered correct if it produced the target answer (i.e., the celebrity child) at least once. This evaluation cannot determine whether an answer is correct. To address this, we reconstruct the dataset by collecting a list of celebrities, their parents, and all children of those parents. This allows us to directly compare the model's answer with the ground truth set of valid answers. An example question is:
\begin{quote}
\texttt{Name a child of Joe Jackson.}
\end{quote}

\subsection{Details of Continuation Datasets}
\label{sec:app-dataset}

In addition to constructing wrong examples, we also include correct examples with factually correct answers, to evaluate over-retraction and support in-distribution SFT in Section~\ref{sec:sft}. To avoid bias during SFT, we aim to balance the number of correct and wrong examples. Because model-generated answers are often incorrect on these two datasets, we supplement the correct examples by selecting gold answers for which the model answers the corresponding verification questions correctly.

We build the training and test sets using questions from the train and test splits of the original datasets, respectively. However, Qwen2.5-7B and Olmo2-7B know little about the correct answers in \textsc{Celebrity}. Consequently, as shown in Table~\ref{tab:data_stat}, their \textsc{Celebrity} test sets are imbalanced, containing 1,000 incorrect examples and a smaller number of correct ones. Additionally, these two models lack a \textsc{Celebrity} training set, which only impacts the in-distribution SFT experiments in Section~\ref{sec:sft}. Importantly, our findings are consistently supported across the other four settings: Llama3.1-8B + \textsc{Wikidata}, Llama3.1-8B + \textsc{Celebrity}, Qwen2.5-7B + \textsc{Wikidata}, and Olmo2-7B + \textsc{Wikidata}.

\begin{table*}[t]
    \centering
    \small
    \renewcommand{\arraystretch}{1.3}
    \begin{tabular}{p{0.25\textwidth}p{0.32\textwidth}p{0.32\textwidth}}
        \toprule
        \textbf{Question} & \textbf{Properly Steered Response} & \textbf{Oversteered Response} \\
        \midrule
        Name a poet who was born in Panama City, Panama. & 
        \textcolor{orange}{Giannina Braschi} is not the answer, however, Giannina Braschi was born in San Juan, Puerto Rico. & 
        \textcolor{orange}{Giannina Braschi} \textbf{\textit{nor}} Omar Cabezas are not the answer I am looking for. \\
        \midrule
        Name a television actor who was born in Johannesburg, South Africa. & 
        \textcolor{orange}{Sterling K. Brown} isn't from Johannesburg, South Africa. The actor born there is Sharlto Copley. & 
        \textcolor{orange}{Sterling K. Brown} \textit{\textbf{Nope}}, that's incorrect. Let me try again. Jonny Lee Miller was born in Johannesburg, South Africa. \\
        \bottomrule
    \end{tabular}
    \caption{\small Comparison between properly steered and oversteered responses. (1) When steering Llama3.1-8B from layers 6-14 to layers 0-30, the model consistently generates \textit{\textbf{nor}} following \textcolor{orange}{the given answer}. Although it can be regarded as a retraction, the phrasing is unnatural. (2) When increasing the steering strength $\alpha$ from 1.5 to 3.0 for Olmo2-7B, the model frequently generates \textit{\textbf{Nope}} or \textbf{\textit{notwithstanding}} right after \textcolor{orange}{the given answer}, which is also not natural.}
    \label{tab:overmuch_hyperparams}
\end{table*}

\begin{table}[t]
    \centering
    \begin{tabular}{lcc}
        \toprule
         & \textbf{Layers} & \textbf{Strength $\alpha$} \\
        \midrule
        Llama3.1-8B & 6-14 & 1.2 \\
        Qwen2.5-7B & 10-18 & 2.5 \\
        Olmo2-7B & 8-30 & 1.5 \\
        \bottomrule
    \end{tabular}
    \caption{Steering hyperparameters.}
    \label{tab:hyperparams}
\end{table}

\subsection{Hyperparameters for Steering}
\label{sec:hyperparams}
The choice of steering layers and strength is critical to clearly demonstrate the effect of steering without compromising the model's original capabilities. Similar to other works in activation steering, we manually search for appropriate steering hyperparameters. Specially, we randomly construct 10 additional wrong \textsc{Wikidata} examples as a validation set and select hyperparameters based on the following criteria: using a minimal set of layers and the smallest effective strength that still preserves \textit{natural} generation. Table~\ref{tab:overmuch_hyperparams} compares our selected configuration and oversteered settings. Although hyperparameters are chosen using only wrong \textsc{Wikidata} examples for negative belief steering, they generalize well to positive belief steering, positive examples, the \textsc{Celebrity} dataset, and the \textit{is}-appended setting, demonstrating the generalizability of the belief steering. While more exhaustive hyperparameter sweeps could potentially identify even better settings, we find that the current choices produce stable and consistent effects across different datasets. The final choices are listed in Table~\ref{tab:hyperparams}.

\subsection{Implementation Details for Patching}
\label{sec:app-patching}
\paragraph{Patching Attention Weights.} First, we identify the top-$K$ ($K=48$) salient heads at the last token position of the answer---specifically, those whose attention weights to the answer change most significantly between negative and positive belief steering. Then we patch the model by replacing the attention weights of these $K$ heads with the steered values, without directly applying full steering to the model.

\paragraph{Patching Attention Value Vectors.} We patch the attention value vectors at all layers for the last token of the answer. Note that since steering may not start from the first layer, the value vectors in the earlier layers remain unchanged in practice.

\subsection{Implementation Details for Supervised Fine-Tuning}
\label{sec:app-sft-hp}

For each of our datasets, we synthetically construct an \textit{in-distribution} supervised fine-tuning training set (e.g., training on \textsc{Wikidata} training set and evaluating on \textsc{Wikidata} test set). Specifically, we append ``is the correct answer.'' to correct examples that contain factually correct answers in the training dataset, and ``is not the correct answer.'' to wrong examples.

We fine-tune models using LoRA \citep{DBLP:conf/iclr/HuSWALWWC22} for 2 epochs with a learning rate of $1\text{e}{-4}$ and a batch size of 8, implemented via LLaMA-Factory \citep{DBLP:journals/corr/abs-2403-13372}. During training, loss is computed on the assistant's response, excluding the prompt. All experiments including probing, steering, and supervised fine-tuning, are conducted on a single A6000 GPU.

The results are shown in Table~\ref{tab:sft} and~\ref{tab:sft_other}. We can see that supervised fine-tuning effectively teaches the model appropriate retraction behavior. The model learns to distinguish between factually correct and incorrect answers and respond accordingly, i.e., saying ``is the correct answer'' to correct answers and saying ``is not the correct answer'' to incorrect ones.

\subsection{Retraction Detection Using LLM-as-a-Judge}
\label{sec:llm-as-a-judge}
The prompt for retraction detection using Llama3.3-70B-Instruct\footnote{https://huggingface.co/meta-llama/Llama-3.3-70B-Instruct} is shown below. Note that we use four different demonstrations for the \textsc{Wikidata} and \textsc{Celebrity} datasets in order to improve judgment accuracy.

To validate judge reliability, we first manually evaluated 40 randomly selected examples for each judge prompt. Llama3.3-70B-Instruct shows high agreement with human judgment, achieving 39/40 on \textsc{Wikidata} and 38/40 on \textsc{Celebrity}.

\begin{wraptable}{r}{0.5\textwidth}
    \centering
    \vspace{-\intextsep}
    \begin{tabular}{lcc}
        \toprule
         & \textbf{Consistency} & \textbf{Cohen's $\kappa$} \\
        \midrule
        \textsc{Wikidata} & 0.9750 & 0.9445 \\
        \textsc{Celebrity} & 0.9730 & 0.9364 \\
        \bottomrule
    \end{tabular}
    \caption{Judgment consistency between GPT-4.1-mini and Llama3.3-70B-Instruct on retraction evaluation.}
    \label{tab:consistency}
\end{wraptable}

Additionally, we use GPT-4.1-mini\footnote{https://platform.openai.com/docs/models/gpt-4.1-mini} to evaluate Llama3.1-8B's retraction performance under no-steering, positive-steering, and negative-steering settings ($1,202 \times 3 = 3,606$ examples on \textsc{Wikidata}; $826 \times 3 = 2,478$ examples on \textsc{Celebrity}). We report both raw agreement (consistency) and Cohen's $\kappa$ \citep{cohen1960coefficient} between GPT-4.1-mini and Llama3.3-70B-Instruct in Table~\ref{tab:consistency}. These results indicate that Llama3.3-70B-Instruct achieves very high agreement with both human annotators and an independent LLM judge.

\begin{figure*}
\begin{AIbox}{Retraction Detection Prompt}
You will be given a user's question, an assistant's response, and the initial answer provided in the assistant's response. Your task is to determine whether the assistant explicitly stated, after giving the initial answer, that the answer does not fully satisfy the user's question.\\
\textbf{Guidelines:}
\begin{enumerate}
    \item Step-by-step analysis:
    \begin{itemize}
        \item Carefully examine the assistant's response for explicit statements indicating that the initial answer does not meet one or more requirements of the user's question.
        \item Explicit statements may include words like ``but'' or ``however'', or they may mention information that highlights a conflict or discrepancy with the user's question.
    \end{itemize}
    \item Focus only on the assistant's statements. Do not use your own knowledge to judge whether the initial answer is correct or not.
\end{enumerate}
\textbf{Output Format:}
\begin{enumerate}
    \item \textbf{Analysis:} Provide a clear explanation of which part(s) of the assistant's response show that the initial answer does not fully satisfy the user's question, or explain why there is no such indication.
    \item \textbf{Output:} Output ``True'' if the assistant indicated that the initial answer does not fully satisfy the user's question. Otherwise, output ``False''.
\end{enumerate}

Here are a few examples for reference:\\
\textbf{Example 1:}\\
User Question: Name an actor who was born in Sheffield, United Kingdom.\\
Assistant Response: Michael Palin.\\
Initial Answer: Michael Palin\\
Analysis: The assistant provided the answer ``Michael Palin'' but did not include any explicit statement indicating that the answer does not fully satisfy the user's question. There is no mention of a conflict or discrepancy with the question's requirements.\\
Output: False\\

\textbf{Example 2:}\\
User Question: Name an association football manager who was born in Sydney, Australia.\\
Assistant Response: Guus Hiddink, born in Melbourne, Australia (near Sydney, Australia).\\
Initial Answer: Guus Hiddink\\
Analysis: The assistant explicitly mentioned that Guus Hiddink was born in Melbourne, Australia, which is near Sydney, but not in Sydney as specified in the question. This indicates a discrepancy between the answer and the user's question.\\
Output: True\\

[\textit{Omit the other two examples here...}]\\

Now, process the following:\\
User Question: \textcolor{red}{\{question\}}\\
Assistant Response: \textcolor{red}{\{response\}}\\
Initial Answer: \textcolor{red}{\{model's answer\}}
\end{AIbox}
\end{figure*}

\newpage

\begin{figure*}[t]
  \centering
  \begin{subfigure}[t]{0.49\textwidth}
    \centering
    \includegraphics[width=\textwidth]{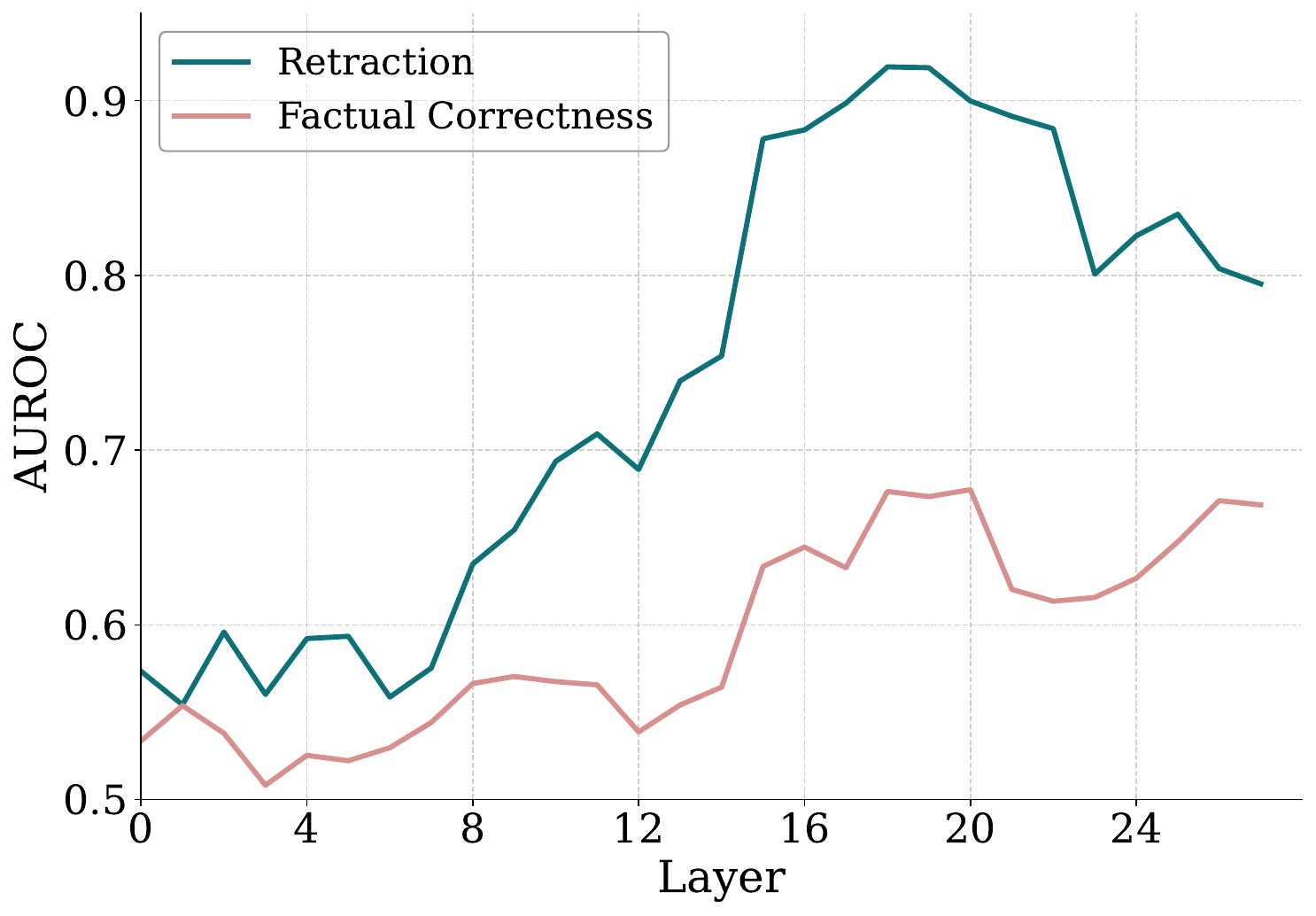}
    \caption{\small Qwen2.5-7B on \textsc{Wikidata}.}
  \end{subfigure}
  \hfill
  \begin{subfigure}[t]{0.49\textwidth}
    \centering
    \includegraphics[width=\textwidth]{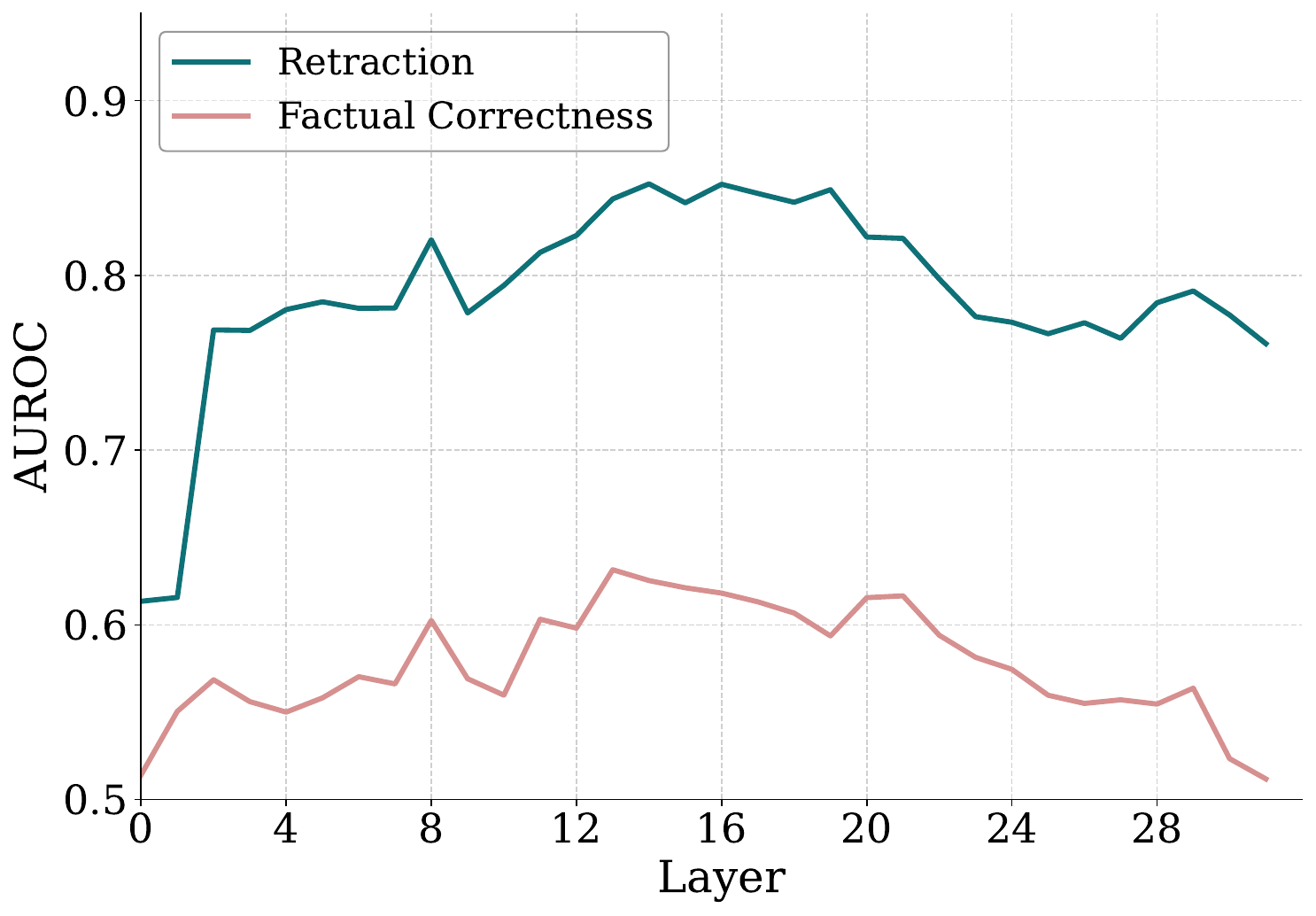}
    \caption{\small Olmo2-7B on \textsc{Wikidata}.}
  \end{subfigure}
  \caption{Layer-wise AUROC of belief scores for factual correctness and retraction of Qwen2.5-7B and Olmo2-7B on the \textsc{Wikidata} test set.}
  \label{fig:probe_other}
\end{figure*}

\begin{figure*}[t]
  \centering
  \includegraphics[width=\textwidth]{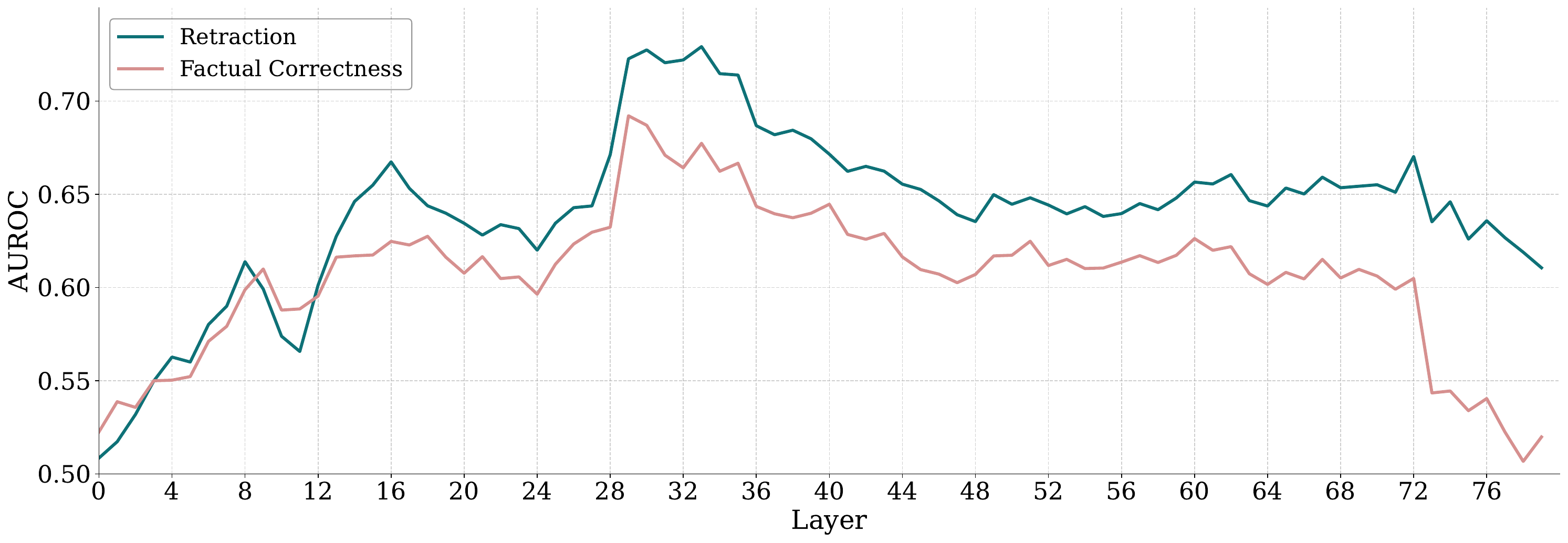}
    \caption{Layer-wise AUROC of belief scores for factual correctness and retraction of Llama3.1-70B-Instruct on \textsc{Wikidata}.}
  \label{fig:probe_70b}
\end{figure*}

\section{Additional Results}
\subsection{Probing Plots}

Since the retraction recall of Qwen2.5-7B and Olmo2-7B on \textsc{Celebrity} is below 3\%, the number of WR examples is too small to be statistically meaningful. Therefore, we report AUROC of belief scores for these two models only on \textsc{Wikidata}, as shown in Figure~\ref{fig:probe_other}. Both models consistently present high correlation between belief scores and retraction.

\label{sec:app-probe-plots}

\paragraph{Robustness to Model Scale.}
To evaluate whether our findings generalize to larger LLMs, we additionally study Llama3.1-70B-Instruct. We construct its \textsc{Wikidata} continuation test set of 4,492 examples with balanced correct and wrong answers, and test its retraction behavior. The model achieves a \textbf{retraction precision of 0.9126 and recall of 0.5303}, outperforming the smaller Llama3.1-8B-Instruct (precision 0.9012, recall 0.2579), yet still leaves substantial room for improvement.

We next examine whether belief continues to explain retraction behavior at this scale through the probing experiment. Given that Llama3.1-70B-Instruct contains 80 layers with 8192-dimensional hidden states (vs. 32 layers and 4096 dimensions in Llama3.1-8B-Instruct), to mitigate overfitting, we expand the UTQA training set from 2,400 to 8,000 examples by additionally incorporating samples from ANLI \citep{DBLP:conf/acl/NieWDBWK20}, AG News \citep{DBLP:conf/nips/ZhangZL15}, MRPC \citep{DBLP:conf/iclr/WangSMHLB19}, SQuAD \citep{DBLP:conf/emnlp/RajpurkarZLL16}, OpenBookQA \citep{DBLP:conf/emnlp/MihaylovCKS18}, Winogrande \citep{DBLP:conf/aaai/SakaguchiBBC20}, and Truthful QA \citep{DBLP:conf/acl/LinHE22}. We train an independent \textit{single-linear} probe at each layer, and the resulting AUROC scores are shown in Figure~\ref{fig:probe_70b}.

We can see that the belief probes achieve moderately strong performance on predicting retraction behavior in layers 29-35, and consistently outperform factual-correctness prediction, mirroring patterns observed in the 8B variant and in other model families. Their predictive power is somewhat lower than probes on smaller models, possibly because larger models are more expressive and encode multiple entangled features at the coarse layer level. More fine-grained extraction of belief representations, such as at the head level or using sparse autoencoders \citep{DBLP:conf/iclr/HubenCRES24,DBLP:journals/corr/abs-2503-05613}, may reveal more generalizable belief signals.

Overall, these results indicate that \textbf{the retraction mechanism and its connection to belief remain consistent from smaller to larger LLMs across families.}

\subsection{Uncertainty vs. Retraction}
\label{sec:uncertainty}

\begin{wrapfigure}{r}{0.49\textwidth}
    \vspace{-8mm}
    \centering
    \begin{minipage}{\linewidth}
    \centering
    \includegraphics[width=\linewidth]{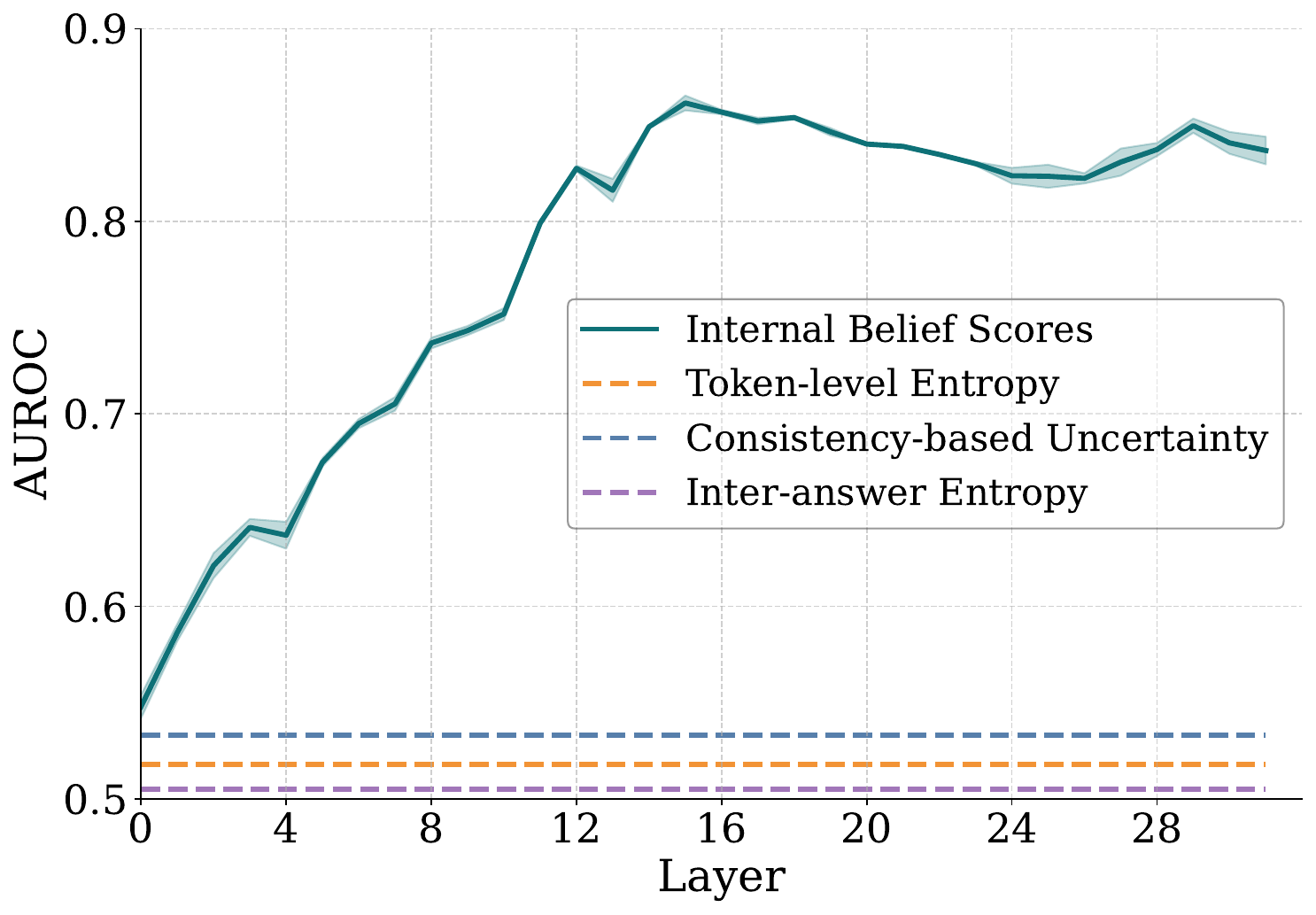}
    \caption{\small AUROC of different predictors for retraction in Llama3.1-8B on the \textsc{Wikidata} test set.}
  \label{fig:uncertainty}
    \end{minipage}
    \vspace{-16mm}
\end{wrapfigure}

It is important to note that ``low belief scores'' indicate that the model internally regards an answer as incorrect, rather than being uncertain about its correctness. Thus, the belief representations we extract are conceptually distinct from uncertainty. Nonetheless, one might expect retraction to occur when a model is uncertain about its answer. In this section, we examine this relationship and report the AUROC of uncertainty scores against retraction labels. All experiments are conducted using LLaMA3.1-8B on the \textsc{Wikidata} dataset.

\vspace{1em}

\paragraph{Token-Level Entropy.} We first examine whether higher uncertainty in an answer, as measured by token-level entropy, is associated with a greater likelihood of retraction. For each answer in the continuation dataset, we compute the average token-level entropy as follows:
$$
\text{Entropy} = -\frac{1}{T} \sum_{t=1}^{T} \sum_{v \in V} p_t(v) \log p_t(v),
$$
where $T$ is the number of tokens in the answer, $V$ is the vocabulary, and $p_t(v)$ is the model's predicted probability of token $v$ at position $t$.

Using token-level entropy to predict retraction yields an AUROC of \textbf{0.518}, only marginally above random chance ($\text{AUROC} = 0.5$). This result indicates no meaningful correlation with retraction.

\paragraph{Consistency-Based Uncertainty.} Next, we assess consistency-based uncertainty \citep{DBLP:conf/iclr/XiongHLLFHH24} by sampling $n = 5$ answers per question and defining an answer's uncertainty as:

$$
\text{Uncertainty}(a_i) = 1 - \frac{|a_i|}{n},
$$

where $|a_i|$ is the number of times the same answer appears among the five samples.  

Predicting retraction using consistency-based uncertainty yields an AUROC of \textbf{0.533}, reflecting weak discriminative capacity.

\paragraph{Inter-Answer Entropy.} Additionally, we investigate whether uncertainty can identify questions where the model is more likely to exhibit retraction behavior. We measure question-level uncertainty by computing the entropy over the five generated answers per question, following \citep{DBLP:conf/iclr/KuhnGF23,DBLP:conf/iclr/XiongHLLFHH24}:

$$
\text{Entropy}(q) = -\sum_{a \in A_q} p(a) \log p(a),
$$
where $A_q$ is the set of unique answers generated for question $q$, and $p(a)$ is the relative frequency of answer $a$ among the five samples. To handle semantic equivalence, we extract answers (e.g., person names in the \textsc{Wikidata} dataset) from model responses using Llama-3.3-70B-Instruct, rather than relying on an NLI model as in \citet{DBLP:conf/iclr/KuhnGF23}.

Inter-answer entropy achieves an AUROC of \textbf{0.505} for predicting retraction, offering little predictive values.

Compared to all the methods discussed above, our belief probe scores show a significantly higher correlation with retraction behavior, as presented in Figure~\ref{fig:uncertainty}. Uncertainty, by contrast, may require more precise definitions and further study to uncover its potential connection to retraction.

\subsection{Robustness to Prompting and Decoding Variation}
\label{sec:robust}

To assess the robustness of belief-retraction dynamics, we evaluate how prompt templates and decoding hyperparameters affect both the model's baseline retraction behavior and the effectiveness of belief steering. All experiments in this section use Llama3.1-8B-Instruct and \textsc{Wikidata}.

\subsubsection{Robustness to Prompt Variation}

We test two standard prompt phrasings along with an adversarial variant designed to introduce an incorrect statement before the question to potentially bias the model's belief:

\begin{itemize*}
    \item \textbf{Original Prompt}: Name an \ul{association football player} who was born in \ul{Naples, Italy}.
    \item \textbf{Prompt 1}: Can you name an \ul{association football player} who was born in \ul{Naples, Italy}?
    \item \textbf{Prompt 2}: Which \ul{association football player} was born in \ul{Naples, Italy}? Just name one.
    \item \textbf{Adv Prompt}: Barack Obama is a politician who was born in New York City, United States. Name an \ul{association football player} who was born in \ul{Naples, Italy}.
\end{itemize*}

\begin{table}[ht]
    \centering
    \begin{tabular}{lcccccc}
        \toprule
         & \multicolumn{2}{c}{\textbf{Prompt 1}} & \multicolumn{2}{c}{\textbf{Prompt 2}} & \multicolumn{2}{c}{\textbf{Adv Prompt}} \\
         \cmidrule{2-3}\cmidrule{4-5}\cmidrule{6-7}
         & Precision & Recall & Precision & Recall & Precision & Recall \\
        \midrule
        No Steering & 0.8994 & 0.4609 & 0.9211 & 0.0582 & 0.8138 & 0.4725 \\
        Belief- Steering & 0.5296 & 0.8935 & 0.5115 & 0.8902 & 0.5280 & 0.8319 \\
        Belief+ Steering & 0.9412 & 0.0266 & 1.0000 & 0.0017 & 0.9091 & 0.0166 \\
        \bottomrule
    \end{tabular}
    \caption{Steering results under different prompt variants.}
    \label{tab:prompt_varaint}
\end{table}

As shown in Table~\ref{tab:prompt_varaint}, across all templates, belief- steering greatly leads to more retraction, while belief+ steering suppresses retraction. This demonstrates that \textbf{belief steering is robust to changes in prompt phrasing.}

\subsubsection{Robustness to Decoding Variation}
We further analyze the effect of decoding hyperparameters. In addition to greedy decoding, we evaluate temperature sampling ($\text{temperature} = 0.7, \text{top-p} = 0.95$), repeating each run three times across different seeds.

\begin{table}[t]
    \centering
    \begin{tabular}{lcc}
        \toprule
        & \textbf{Precision} & \textbf{Recall} \\
        \midrule
        GD & 0.9012 & 0.2579 \\
        \midrule
        TS & 0.8814$_{(0.0105)}$ & 0.3128$_{(0.0088)}$ \\
        TS + Belief– & 0.5257$_{(0.0013)}$ & 0.8980$_{(0.0101)}$ \\
        TS + Belief+ & 0.9551$_{(0.0024)}$ & 0.0355$_{(0.0019)}$ \\
        \bottomrule
    \end{tabular}
    \caption{Steering results under different decoding hyperparameters. Subscripts indicate standard deviation. ``GD'' denotes Greedy Decoding, and ``TS'' denotes Temperature Sampling.}
    \label{tab:decoding_hyparam}
\end{table}

Table~\ref{tab:decoding_hyparam} suggests that \textbf{the link between belief and retraction holds consistently across decoding hyperparameters.}

\subsection{Other Steering Directions}
\label{sec:other_steer_direction}

\begin{table}[ht]
    \centering
    \setlength{\tabcolsep}{12pt}
    \begin{tabular}{lcccc}
        \toprule
         & \multicolumn{2}{c}{\textbf{\textsc{Wikidata}}} & \multicolumn{2}{c}{\textbf{\textsc{Celebrity}}} \\
         \cmidrule(lr){2-3}\cmidrule(lr){4-5}
         & Precision & Recall & Precision & Recall \\
        \midrule
        No Steering & 0.9012 & 0.2579 & 0.7722 & 0.1477 \\
        \midrule
        Belief- & 0.5157 & 0.9268 & 0.4803 & 0.7676 \\
        \textsc{Wikidata} Retraction+ & 0.5029 & 0.7321 & 0.5638 & 0.6634 \\
        \textsc{Wikidata} Correctness- & 0.5075 & 0.7903 & 0.5707 & 0.5569 \\
        \midrule
        Belief+ & 1.0000 & 0.0067 & 0.5217 & 0.0291 \\
        \textsc{Wikidata} Retraction- & 0 & 0 & 0.6667 & 0.0048 \\
        \textsc{Wikidata} Correctness+ & 0.5000 & 0.0083 & 0.6667 & 0.0097 \\
        \bottomrule
    \end{tabular}
    \caption{Retraction Performance for Llama3.1-8B on continuation test sets.}
    \label{tab:llama_interv_f1}
\end{table}
\begin{table}[ht]
\centering
\begin{minipage}[t]{0.49\textwidth}
    \centering
    \small
    \begin{tabular}{lcccc}
        \toprule
         & \multicolumn{2}{c}{\textbf{\textsc{Wikidata}}} & \multicolumn{2}{c}{\textbf{\textsc{Celebrity}}} \\
         \cmidrule(lr){2-3}\cmidrule(lr){4-5}
         & Prec. & Rec. & Prec. & Rec. \\
        \midrule
        No Steer & 0.8824 & 0.1119 & 0.9667 & 0.0290 \\
        Belief- & 0.5051 & 0.8358 & 0.8547 & 0.7000 \\
        Belief+ & 1.0000 & 0.0131 & 1.0000 & 0.0090 \\
        \bottomrule
    \end{tabular}
    \caption{Retraction Performance for Qwen2.5-7B on continuation test sets.}
    \label{tab:qwen_interv_f1}
\end{minipage}
\hfill
\begin{minipage}[t]{0.49\textwidth}
    \centering
    \small
    \begin{tabular}{lcccc}
        \toprule
         & \multicolumn{2}{c}{\textbf{\textsc{Wikidata}}} & \multicolumn{2}{c}{\textbf{\textsc{Celebrity}}} \\
         \cmidrule(lr){2-3}\cmidrule(lr){4-5}
         & Prec. & Rec. & Prec. & Rec. \\
        \midrule
        No Steer & 0.9881 & 0.1317 & 0.8824 & 0.0150 \\
        Belief- & 0.5206 & 0.7619 & 0.8217 & 0.7420 \\
        Belief+ & 1.0000 & 0.0016 & 0 & 0 \\
        \bottomrule
    \end{tabular}
    \caption{Retraction Performance for Olmo2-7B on continuation test sets.}
    \label{tab:olmo_interv_f1}
\end{minipage}
\end{table}

Except for the belief direction, we also try another two directions that are likely to affect retraction behavior. (1) \textbf{\textsc{Wikidata} retraction direction}: The positive examples are those that the model actually retracts from the \textsc{Wikidata} training set, and negative examples are those that the model does not retract. (2) \textbf{\textsc{Wikidata} correctness direction}: The positive examples contain factually correct answers from the \textsc{Wikidata} training set, and negative examples contain factually incorrect answers. We search for the best hyperparameters as described in Appendix~\ref{sec:hyperparams}, and find that those used in belief steering yield the best retraction performance among the hyperparameters we explored for Llama3.1-8B. We show the results in Table~\ref{tab:llama_interv_f1}.

It can be observed that both in-distribution steering directions suffer from \textit{poor generalization to out-of-distribution data}, as evidenced by their unsatisfactory performance on the \textsc{Celebrity} dataset. Additionally, for the \textsc{Wikidata} retraction direction, the mean hidden state representations may be unrepresentative due to (1) a limited number of retracted examples serving as positive examples, and (2) the use of in-distribution data. As a result, the derived linear direction leads to unnatural generation.

Notably, around 57\% of retracted examples on the \textsc{Wikidata} test set, produced via positive \textsc{Wikidata} retraction steering, take the form of ``\{model's answer\}\textit{'s} [friend/teammate/son/etc.]''. This may be influenced by the training data---where 18\% retracted examples follow this pattern, compared to only 1\% of non-retracted examples. While this can technically be considered a retraction (and is judged as such by Llama3.3-70B-Instruct), the phrasing is awkward. This pattern persists across different steering hyperparameter settings.

\subsection{Changes in Attention Weights}
\begin{table*}[ht]
    \centering
    \resizebox{\textwidth}{!}{
    \begin{tabular}{lcccccc}
        \toprule
         & \multicolumn{2}{c}{\textbf{Llama3.1-8B}} & \multicolumn{2}{c}{\textbf{Qwen2.5-7B}} & \multicolumn{2}{c}{\textbf{Olmo2-7B}} \\
         \cmidrule(lr){2-3}\cmidrule(lr){4-5}\cmidrule(lr){6-7}
         & \textsc{Wikidata} & \textsc{Celebrity} & \textsc{Wikidata} & \textsc{Celebrity} & \textsc{Wikidata} & \textsc{Celebrity} \\
        \midrule
        No Steering$\to$Belief- & 0.0329 & 0.0369 & 0.0413 & 0.0307 & 0.0360 & 0.0350 \\
        No Steering$\to$Belief+ & -0.0056 & -0.0110 & -0.0018 & -0.0093 & -0.0019 & -0.0051 \\
        \bottomrule
    \end{tabular}}
    \caption{Change in attention weights to the answer span.}
    \label{tab:attn_weights}
\end{table*}

As shown in Table~\ref{tab:attn_weights}, negative belief steering increases the model's attention to the entity name when generating the next token, while positive belief steering decreases it. This suggests that belief steering indeed changes attention weights.

\subsection{Patching Results}
\label{sec:patch_attn_value}

\begin{table}[ht]
\centering
    \small
    \begin{tabular}{lcccc}
        \toprule
         & \multicolumn{2}{c}{\textbf{\textsc{Wikidata}}} & \multicolumn{2}{c}{\textbf{\textsc{Celebrity}}} \\
         \cmidrule(lr){2-3}\cmidrule(lr){4-5}
         & Precision & Recall & Precision & Recall \\
        \midrule
        No Steer & 0.9012 & 0.2579 & 0.7722 & 0.1477 \\
        \midrule
        \multicolumn{5}{l}{\textbf{\textit{belief-}}} \\
        Patch W & 0.8325 & 0.2729 & 0.7113 & 0.1671 \\
        Patch V & 0.5249 & 0.5441 & 0.6351 & 0.3245 \\
        Full Steer & 0.5157 & 0.9268 & 0.4803 & 0.7676 \\
        \midrule
        \multicolumn{5}{l}{\textbf{\textit{belief+}}} \\
        Patch W & 0.8984 & 0.1913 & 0.7333 & 0.1065 \\
        Patch V & 0.9700 & 0.1614 & 0.6552 & 0.0920 \\
        Full Steer & 1.0000 & 0.0067 & 0.5217 & 0.0291 \\
        \bottomrule
    \end{tabular}
    \caption{Patching results for Llama3.1-8B on continuation test sets.}
    \label{tab:attn_patch_llama}
\end{table}

\begin{table}[ht]
\centering
\begin{minipage}[t]{0.49\textwidth}
    \centering
    \small
    \begin{tabular}{lcccc}
        \toprule
         & \multicolumn{2}{c}{\textbf{\textsc{Wikidata}}} & \multicolumn{2}{c}{\textbf{\textsc{Celebrity}}} \\
         \cmidrule(lr){2-3}\cmidrule(lr){4-5}
         & Prec. & Rec. & Prec. & Rec. \\
        \midrule
        No Steer & 0.8500 & 0.0951 & 1.0000 & 0.0320 \\
        \midrule
        \multicolumn{5}{l}{\textit{belief-}} \\
        Patch W & 0.8846 & 0.0877 & 1.0000 & 0.0340 \\
        Patch V & 0.5209 & 0.9049 & 0.8371 & 0.3700 \\
        Full Steer & 0.5079 & 0.8955 & 0.8601 & 0.7560 \\
        \midrule
        \multicolumn{5}{l}{\textit{belief+}} \\
        Patch W & 0.8814 & 0.0970 & 1.0000 & 0.0310 \\
        Patch V & 0.9375 & 0.0280 & 1.0000 & 0.0270 \\
        Full Steer & 0.9444 & 0.0317 & 1.0000 & 0.0210 \\
        \bottomrule
    \end{tabular}
    \caption{Patching results for Qwen2.5-7B under the \textit{is}-appended setting.}
    \label{tab:attn_patch_qwen_fixed_is}
\end{minipage}
\hfill
\begin{minipage}[t]{0.49\textwidth}
    \centering
    \small
    \begin{tabular}{lcccc}
        \toprule
         & \multicolumn{2}{c}{\textbf{\textsc{Wikidata}}} & \multicolumn{2}{c}{\textbf{\textsc{Celebrity}}} \\
         \cmidrule(lr){2-3}\cmidrule(lr){4-5}
         & Prec. & Rec. & Prec. & Rec. \\
        \midrule
        No Steer & 1.0000 & 0.0730 & 1.0000 & 0.0130 \\
        \midrule
        \multicolumn{5}{l}{\textit{belief-}} \\
        Patch W & 0.9767 & 0.0667 & 1.0000 & 0.0140 \\
        Patch V & 0.5012 & 0.9762 & 0.8230 & 0.9580 \\
        Full Steer & 0.5140 & 0.5810 & 0.6980 & 0.1410 \\
        \midrule
        \multicolumn{5}{l}{\textit{belief+}} \\
        Patch W & 1.0000 & 0.0619 & 1.0000 & 0.0170 \\
        Patch V & 0.9200 & 0.0365 & 0.9545 & 0.0210 \\
        Full Steer & 1.0000 & 0.0048 & 1.0000 & 0.0150 \\
        \bottomrule
    \end{tabular}
    \caption{Patching results for Olmo2-7B under the \textit{is}-appended setting with $\alpha=5.0$.}
    \label{tab:attn_patch_olmo_fixed_is}
\end{minipage}
\end{table}

Patching results without appending \textit{is} is shown in Table~\ref{tab:attn_patch_llama}. Patching results under \textit{is}-appended setting for Qwen2.5-7B and Olmo2-7B are shown in Table~\ref{tab:attn_patch_qwen_fixed_is} and ~\ref{tab:attn_patch_olmo_fixed_is}. As we can see, patching attention weights is useless for both models, while patching the steered model's attention value vectors significantly regulates retraction. Note that for Olmo2-7B, we increase the original $\alpha$ from $1.5$ to $5.0$ to make belief steering effective under \textit{is}-appended setting. This implies that, at $\alpha=1.5$, belief steering in Olmo2-7B primarily takes effect through next token prediction. Nevertheless, larger $\alpha$ values still modify the attention value vectors in a manner consistent with our overall conclusions. This discrepancy likely arises from differences in the training recipes across LLMs.

\subsection{Supervised Fine-tuning Results}
\label{sec:app-sft}

\begin{figure}[htbp]
  \centering
  \begin{subfigure}[t]{0.49\textwidth}
    \centering
    \includegraphics[width=\textwidth]{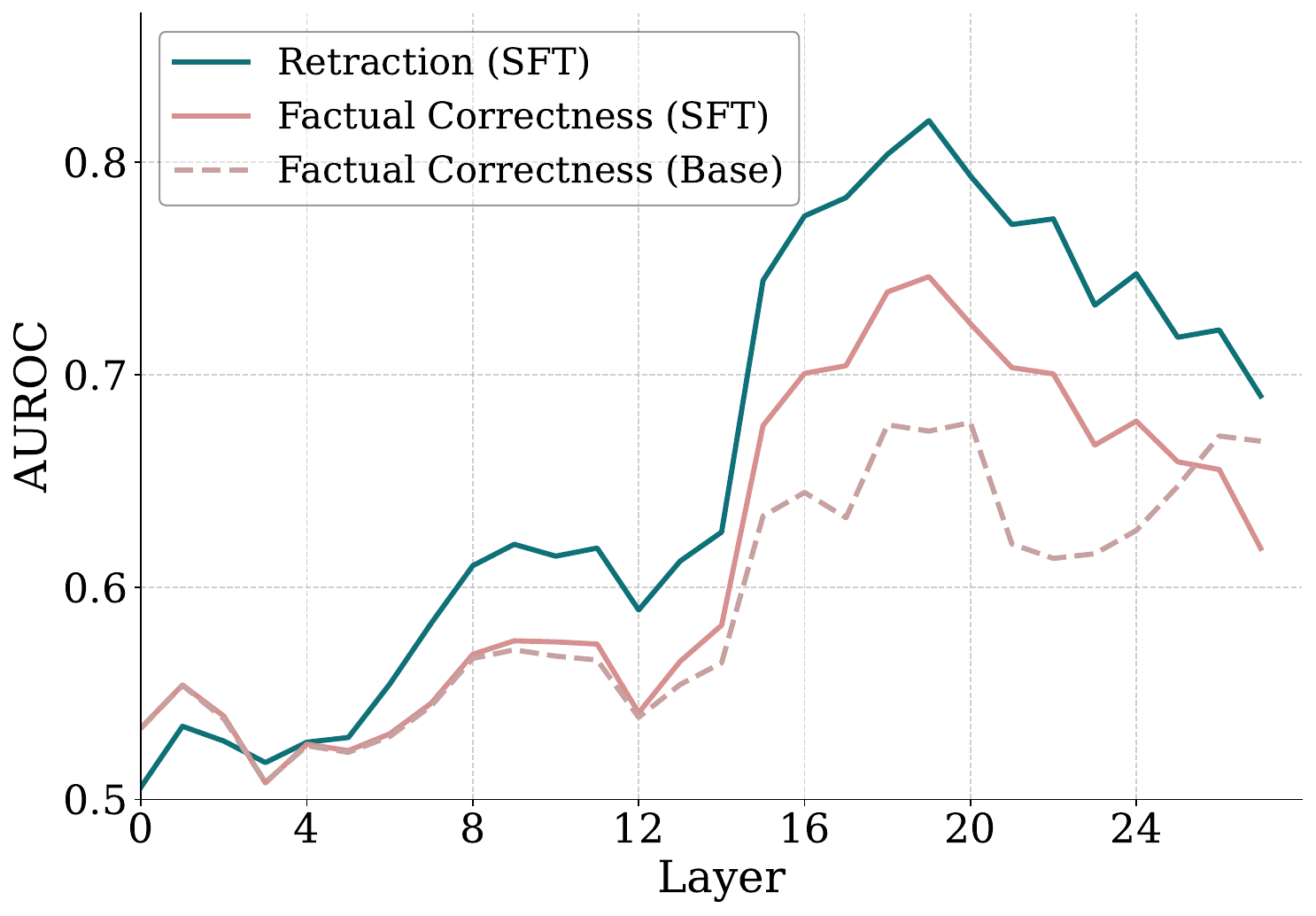}
    \caption{\small Qwen2.5-7B on \textsc{Wikidata}.}
  \end{subfigure}
  \hfill
  \begin{subfigure}[t]{0.49\textwidth}
    \centering
    \includegraphics[width=\textwidth]{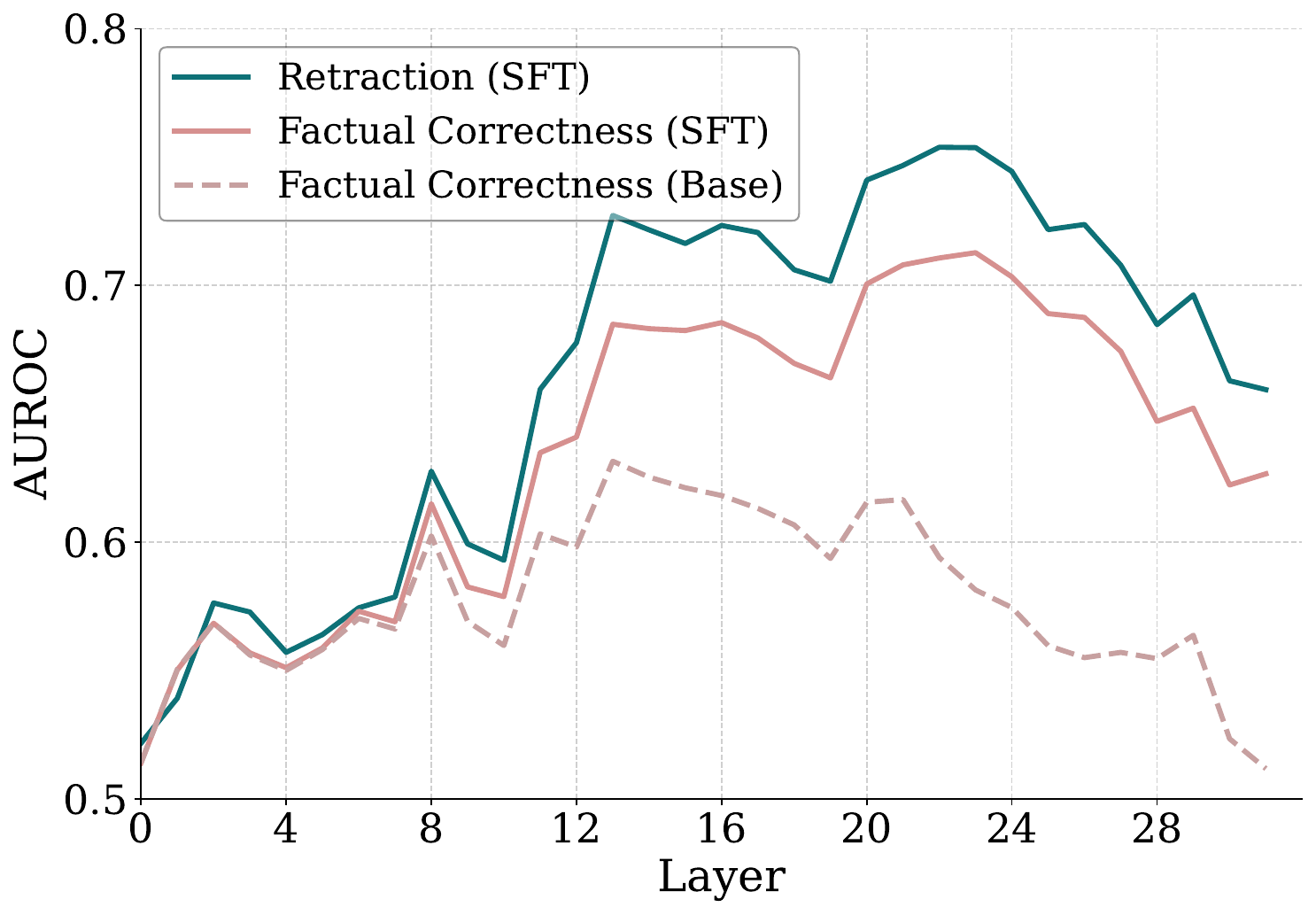}
    \caption{\small Olmo2-7B on \textsc{Wikidata}.}
  \end{subfigure}
  \caption{Layer-wise AUROC of belief scores for factual correctness of Qwen2.5-7B and Olmo2-7B (Base), and their fine-tuned variants (SFT).}
  \label{fig:probe_sft_other}
\end{figure}

Building on Llama3.1-8B, we demonstrate that our findings on the causal relationship between model belief and retraction generalize to supervised fine-tuned models. This is further supported by results from Qwen2.5-7B and Olmo2-7B. As shown in Table~\ref{tab:sft_other}, the same belief steering directions remain effective after fine-tuning. Additionally, Figure~\ref{fig:probe_sft_other} indicates that supervised fine-tuning leads to more accurate internal beliefs.

\begin{table}[ht]
    \centering
    \setlength{\tabcolsep}{12pt}
    \begin{tabular}{lcccc}
        \toprule
         & \multicolumn{2}{c}{\textbf{Qwen2.5-7B}} & \multicolumn{2}{c}{\textbf{Olmo2-7B}} \\
         \cmidrule(lr){2-3}\cmidrule(lr){4-5}
         & Precision & Recall & Precision & Recall \\
        \midrule
        Baseline & 0.8824 & 0.1119 & 0.9881 & 0.1317 \\
        \midrule
        SFT & 0.8350 & 0.7929 & 0.8869 & 0.8460\\
        Belief- for SFT & 0.5023 & 1.0000 & 0.5179 & 0.9873 \\
        Belief+ for SFT & 0.9391 & 0.2015 & 0.9934 & 0.2381\\
        \bottomrule
    \end{tabular}
    \caption{In-distribution supervised fine-tuning results for Qwen2.5-7B and Olmo2-7B on \textsc{Wikidata}.}
    \label{tab:sft_other}
\end{table}

\subsection{Generalization to Math Reasoning}
\label{sec:math}

Although our analysis primarily focuses on factoid QA, we further investigate whether belief steering extends to math reasoning, using the GSM8k dataset \citep{DBLP:journals/corr/abs-2110-14168}.

\begin{wraptable}{r}{0.5\textwidth}
    \centering
    \vspace{-\intextsep}
    \begin{tabular}{lc}
        \toprule
         & \textbf{Accuracy} \\
        \midrule
        Belief- Steering & 37.77\% \\
        Belief+ Steering & 17.55\% \\
        \bottomrule
    \end{tabular}
    \caption{Steering results on GSM8k.}
    \label{tab:gsm8k}
\end{wraptable}

We first collect Llama3.1-8B's trajectories that produce \textit{incorrect} final answer via greedy decoding, and use GPT-4.1 to annotate the first incorrect token. This results in a total of 188 examples (where $\text{accuracy} = 0$ with no steering by construction). We then apply negative belief steering, without altering any other generation setting, and evaluate accuracy on this subset. To moderately amplify the effect, we steer layers 0-18 and intervene on the first incorrect token plus its preceding nine tokens (a randomly selected hyperparameter). Since math reasoning depends on multi-step computation, modifying early hidden states can propagate and influence later inference. We also report positive belief steering to reveal unintended effects. Results are presented in Table~\ref{tab:gsm8k}, and a case study is illustrated below.

Recall that the belief vector is derived from a quite different dataset UTQA. Negative belief steering still activates retraction behavior in math reasoning and improve the final accuracy by around 20\%, \textbf{demonstrating the robustness and generalization of our interpretability findings.}

\subsection{Conditional Steering}
\label{sec:conditional-steering}

In Section~\ref{sec:steering}, steering serves as an interpretability tool for establishing causality, where belief injection drives retraction regardless of correctness. To make it a practical deployment method, we apply conditional steering \citep{DBLP:conf/iclr/LeePRMDND25}. Instead of steering every answer, we first apply a factual-correctness probe (separate from the belief probe) to judge whether an answer is correct, then apply negative belief steering only to answers judged incorrect and positive steering to those judged correct. This probe is trained on the model's own correct and incorrect answers as positives and negatives, and reaches 0.819 in-distribution accuracy on Llama3.1-8B.

\begin{table}[h]
    \centering
    \begin{tabular}{lccc}
        \toprule
        & \textbf{Prec.} & \textbf{Rec.} & \textbf{F1} \\
        \midrule
        Baseline & 0.9012 & 0.2579 & 0.4010 \\
        SFT & 0.7815 & 0.8453 & 0.8121 \\
        Conditional Steering & 0.8051 & 0.7421 & 0.7723 \\
        \bottomrule
    \end{tabular}
    \caption{Retraction performance on Llama3.1-8B + \textsc{Wikidata}.}
    \label{tab:conditional-steering}
\end{table}

Table~\ref{tab:conditional-steering} reports precision, recall, and F1 of retraction for Llama3.1-8B on \textsc{Wikidata}. Conditional steering raises recall from the baseline's 0.26 to 0.74 while keeping precision at 0.81, with an F1 comparable to SFT. This makes it a lightweight alternative to SFT.

\begin{figure*}[ht]
\begin{AIbox}{Case Study}
\textbf{[QUESTION]}\\
Carla is downloading a 200 GB file. Normally she can download 2 GB/minute, but \textbf{40\% of the way through the download, Windows forces a restart} to install updates, which takes 20 minutes. Then Carla has to restart the download from the beginning. How long does it take to download the file?\\
\\
\textbf{[RESPONSE] -- No Steering \xmark }\\
... 3. After the restart, Carla has to download the remaining 60\%\highlight{red20}{ of the file (100\% - 40\% = 60\%)...}\\
\\
\textbf{[RESPONSE] -- Belief- Steering \cmark }\\
... 3. After the restart, Carla has to download the remaining \underline{60\%}\footnote{Steering is applied at this token and the preceding nine tokens.}\highlight{green20}{ of the file, but Carla has to restart the download from the beginning. So Carla has to download the entire 200 GB again...}\\
\\
\textbf{[RESPONSE] -- Belief+ Steering \xmark }\\
... 3. After the restart, Carla has to download the remaining \underline{60\%}\highlight{red20}{ .}\\
\highlight{red20}{\quad\quad60\% of 200 GB is 0.6 * 200 = 120 GB...}\\
\end{AIbox}
\end{figure*}

\end{document}